\documentclass[letterpaper, times]{elsarticle}
\makeatletter
\def\ps@pprintTitle{%
 \let\@oddhead\@empty
 \let\@evenhead\@empty
 \def\@oddfoot{\thepage\hfill\footnotesize\itshape\today}
 \let\@evenfoot\@oddfoot}
\makeatother

\usepackage[margin=2.54cm]{geometry}
\usepackage{xcolor}
\usepackage{mathtools}
\usepackage{graphicx}
\usepackage{amssymb}
\usepackage{caption}
\usepackage{subcaption}
\usepackage{tcolorbox}
\usepackage{scrextend}
\usepackage{listings}
\usepackage{bm}
\usepackage{float}
\usepackage[colorlinks]{hyperref}
\usepackage{multirow}
\usepackage[linesnumbered,lined,boxed,ruled,vlined]{algorithm2e}
\usepackage{afterpage}
\usepackage{etoolbox}
\usepackage{dutchcal}
\usepackage{makecell}
\usepackage{lineno}
\usepackage{soul}
\usepackage[pages=some,placement=top]{background}

\backgroundsetup{
	scale=1,
	color=black,
	opacity=1.0,
	angle=0,
	contents={%
		\includegraphics[width=\paperwidth,height=2cm]{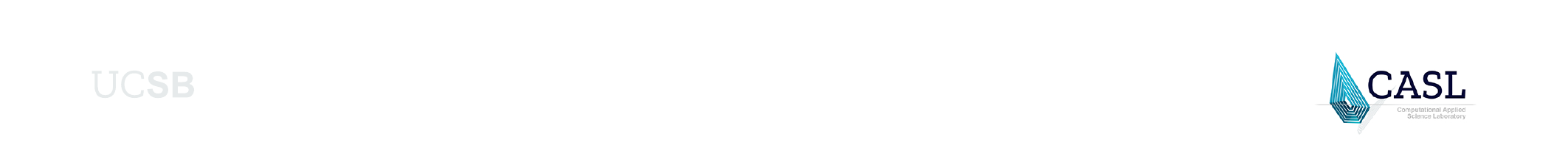}
	}%
}

\hypersetup{
	colorlinks=true,      		
	linkcolor=blue,				
	citecolor=blue,				
	filecolor=black,      		
	urlcolor=red,				
	bookmarks=false,
	pdffitwindow=true,
	pdfpagelayout=SinglePage
}

\definecolor{navy}{RGB}{0, 54, 96}
\definecolor{aqua}{RGB}{4, 124, 145}
\definecolor{darkcoral}{RGB}{196, 52, 36}
\definecolor{mist}{RGB}{156, 190, 190}

\newcommand{\colorsection}[1]{\sffamily\color{aqua}\section{#1}\color{black}\rmfamily}
\newcommand{\colorsubsection}[1]{\sffamily\color{aqua}\subsection{#1}\color{black}\rmfamily}

\patchcmd{\abstract}{Abstract}{\sffamily\textcolor{aqua}{Abstract}\rmfamily}{}{}
\abstracttitle{\sffamily\textcolor{aqua}{Abstract}}

\newcommand{\etal}{et al.\ }

\newcommand{\vv}[1]{ \boldsymbol{\mathbf{#1}} }

\newcommand{\eten}[2]{ #1\times 10^{#2} }

\let\OLDthebibliography\thebibliography
\renewcommand\thebibliography[1]{
  \OLDthebibliography{#1}
  \setlength{\parskip}{1pt}
  \setlength{\itemsep}{1pt plus 0.3ex}
}

\begin{document}

\title{\textcolor{navy}{\sffamily\bfseries A Hybrid Inference System for Improved Curvature Estimation in the Level-Set Method Using Machine Learning}}

\author[1]{\color{navy}Luis \'{A}ngel Larios-C\'{a}rdenas\corref{cor1}}
\author[1,2]{\color{navy}Fr\'{e}d\'{e}ric Gibou}

\cortext[cor1]{Corresponding author: lal@cs.ucsb.edu}

\address[1]{Department of Computer Science, University of California, Santa Barbara, CA 93106, USA}
\address[2]{Department of Mechanical Engineering, University of California, Santa Barbara, CA 93106, USA\color{black}}

\BgThispage


\begin{abstract}

We present a novel hybrid strategy based on machine learning to improve curvature estimation in the level-set method.  The proposed inference system couples enhanced neural networks with standard numerical schemes to compute curvature more accurately.  The core of our hybrid framework is a switching mechanism that relies on well established numerical techniques to gauge curvature.  If the curvature magnitude is larger than a resolution-dependent threshold, it uses a neural network to yield a better approximation.  Our networks are multilayer perceptrons fitted to synthetic data sets composed of sinusoidal- and circular-interface samples at various configurations.  To reduce data set size and training complexity, we leverage the problem's characteristic symmetry and build our models on just half of the curvature spectrum.  These savings lead to a powerful inference system able to outperform any of its numerical or neural component alone.  Experiments with stationary, smooth interfaces show that our hybrid solver is notably superior to conventional numerical methods in coarse grids and along steep interface regions.  Compared to prior research, we have observed outstanding gains in precision after training the regression model with data pairs from more than a single interface type and transforming data with specialized input preprocessing.  In particular, our findings confirm that machine learning is a promising venue for reducing or removing mass loss in the level-set method.

\end{abstract}


\maketitle



\colorsection{Introduction}
\label{sec:Introduction}

Mean curvature is a fundamental interface attribute in free boundary problems \cite{Friedman10} for its relation to surface tension in physics \cite{Popinet;NumModelsOfSurfTension;18} and its regularization property in optimization.  It plays a crucial role in the solution of numerical models for applications such as free surface and multiphase flows \cite{Sussman;Smereka;Osher:94:A-Level-Set-Approach, Sussman;Fatemi;Smereka;etal:98:An-Improved-Level-Se, Gibou;Chen;Nguyen;etal:07:A-level-set-based-sh, Theillard:2019aa, Losasso;Gibou;Fedkiw:04:Simulating-Water-and, Losasso:2006aa, Gibou:2019aa}, image segmentation \cite{Chan;Vese:01:Active-Contour-Witho, Gibou:2005aa}, diffusion-dominated Stefan models \cite{Theillard;Gibou;Pollock:14:A-Sharp-Computationa, Papac;Helgadottir;Ratsch;etal:13:A-level-set-approach,Papac;Gibou;Ratsch:10:Efficient-symmetric-, Chen;Min;Gibou:09:A-numerical-scheme-f, Mirzadeh;Gibou:14:A-conservative-discr}, and morphogenesis \cite{AliasBuenzli20}.

One of the most successful numerical schemes for advecting interfaces is the level-set method \cite{Osher1988, Sethian:99:Level-set-methods-an, GFO18}.  This framework belongs to a family of Eulerian formulations, alongside the volume-of-fluid \cite{Hirt;Nichols:81:Volume-of-Fluid-VOF-} and the phase-field models \cite{Fix:1983aa, Langer:1986aa}.  These methods use an implicit function to both capture and evolve the interface.  The main advantage of Eulerian formulations is their natural ability to handle complex changes in the interface topology.  In particular, implicit methods do not need to resort to complicated procedures to manipulate free boundaries explicitly.  For a short review of these and other families of explicit schemes, we refer the reader to our opening section in \cite{LALariosFGibou;LSCurvatureML;2021}.

Because of its implicit representation, the level-set method provides a simple relation for computing mean curvature, whose accuracy depends on the smoothness of the level-set function \cite{Chene;Min;Gibou:08:Second-order-accurat, Popinet;NumModelsOfSurfTension;18}.  We can enforce such a smoothness by reinitializing the underlying scalar field into a signed distance function with high-order iterative procedures.  However, these reinitialization schemes do not always deliver satisfactory results, especially in under-resolved regions and nonuniform grids.  Indeed, when the interface is well-resolved, level-set approaches can produce very accurate curvature values \cite{Chene;Min;Gibou:08:Second-order-accurat}.  Likewise, there are sophisticated front-tracking implementations (e.g., \cite{iPAM;Zhang-Fogelson;2014, MARS-iPAM;Zhang-Fogelson;2016}) that one can use to evolve interfaces and approximate height functions with a high precision that eventually leads to curvature computations with unparalleled accuracy (e.g., \cite{Zhang;HFES;2017}).  Yet, level-set methods are better suited for extreme topological changes \cite{Osher;Fedkiw:02:Level-Set-Methods-an}, and here we are focusing on interface regions where standard numerical schemes perform poorly.  Our goal is to explore a hybrid strategy that combines inexpensive reinitialization with more recent data-driven technologies.

The last two decades have witnessed a resurgence of machine learning research \cite{A18, Mehta19, Hands-onMLwithScikit-LearnKerasAndTF19} in areas such as image processing (e.g., segmentation \cite{NLC17-Ventricle-Segmentation, LevelSetAsDeepRNN18}, classification \cite{AlexNet12, ResNet16},  reconstruction \cite{A18, SDenoisingAutoEncoders10}, and synthesis \cite{GAdversarialNets14}), natural language and information retrieval, (e.g., distributed word representations \cite{Word2Vec13, BGJM17-Word-Vectors-with-Subword-Info, Elmo18} and context embeddings \cite{ALM17-Sentence-Embeddings}), and sequence and advanced language models (e.g., machine language translation \cite{GRU14, Transformer17}).  Data availability and widespread access to powerful computing resources have fueled a renewed enthusiasm for this technology in the computer science community and other disciplines like mathematics, physics, and engineering \cite{IntroToNeuralMethodsForDiffEqns15}.

Artificial neural networks are computational graphs of elementary units interconnected in some particular way to compute a function of its inputs by using connection weights as intermediate parameters \cite{A18}.  Supervised neural networks are function approximators \cite{Mehta19} capable of capturing a high degree of complexity and nonlinearity.  These models learn by successively adapting their weights to minimize a loss metric given their current parameters and a training data set.  Although training neural networks is a long and expensive process involving backpropagation \cite{A18}, one does this offline and once for each application.  Later, one can use the neural network as a black box to estimate the output for data samples never seen during the learning stage.

Novel research in computational science has incorporated the neural networks' ability to approximate any function. We can trace the use of regression models in this discipline back to the pioneering work of Lagaris \etal in \cite{NNetForODEsPDEs98, NNetForPDEs00}.  Lagaris and coauthors trained their multilayer perceptrons to solve initial and boundary problems in ordinary and partial differential equations.  More recently, Raissi \etal \cite{Raissi17a, Raissi18} have presented physics-informed deep learning and have constructed neural models to discover dynamic dependencies in spatiotemporal data sets.

Over the past few years, practitioners have also integrated neural networks to the numerical solution of conservation laws as troubled-cell indicators \cite{TroubledCellIndicator18}.  These models flag near-discontinuity regions where Runge--Kutta discontinuous Galerkin schemes need to be carefully corrected.  In that case, robust binary classifiers ensure the numerical methods' cost-efficiency and remove the prescription of any problem-dependent parameters.  Similarly, Morgan \etal \cite{ShockDetector20} have trained a shock detector coupled with a high-order Lagrangian hydrodynamic method in complex flows.  The idea is to use a first-order solution in cells lying near a shock and a high-order solution everywhere else in the mesh. The intention is to avoid treating smooth flow features as shocks, which could lead to reduced accuracy.

Lastly, computational fluid dynamics researchers have started to inspect machine learning to address challenging tasks in the volume-of-fluid method (VOF).  In \cite{DespresJourdren;MLDesignOfVOF;20}, Despr\'{e}s and Jourdren have proposed a family of VOF machine learning algorithms suitable to bimaterial compressible Euler calculations.  Their development includes training a five-layered network with lines, arcs, and corners.  Despr\'{e}s and Jourdren have implemented their regression model in a finite volume Lagrange+remap solver, aiming at an accurate transport/remap of reconstructed interfaces.  In like manner, Qi \etal \cite{CurvatureML19} have pioneered a machine learning strategy to reduce the complexity and improve the accuracy of interface curvature calculations in two-dimensional VOF schemes.  In \cite{CurvatureML19}, a neural network ingests (nine-cell stencil) volume fractions and produces continuous dimensionless curvature values for center cells.  To train their neural model, Qi and coauthors used large synthetic data sets, spanning a wide range of curvatures from circular interfaces.  A more recent and detailed work by Patel \etal \cite{VOFCurvature3DML19} has addressed local curvature estimation in three-dimensional VOF schemes.  Their approach has proven effective in standard test cases with multiple bubble simulations.  Patel and colleagues have shown that their solver outperforms the convolution method and even matches the precision of the height function method.  In a parallel VOF study, Ataei \etal \cite{NPLIC20} have proposed a multilayer perceptron to carry out piece-wise linear interface construction (PLIC).  Their approach has reduced computational costs and surpassed the quality of the analytical and iterative traditional PLIC when approximating the location of moving boundaries.

The present research is motivated by the recent progress in \cite{CurvatureML19, VOFCurvature3DML19} and substantially improves the strategy introduced in \cite{LALariosFGibou;LSCurvatureML;2021} to compute curvature in the level-set method.  Our work aligns in spirit with the cost-efficiency goal of \cite{ShockDetector20} too; we seek to use machine learning solvers only when deemed necessary.  The proposed hybrid inference system couples a regression model with standard numerical schemes to estimate curvature more accurately for two-dimensional free boundaries.  We fit our neural networks to synthetic data sets composed of sinusoidal- and circular-interface samples at various configurations.  Further, to reduce data set size and training complexity, we exploit the problem's characteristic symmetry and build our models on just half of the curvature spectrum.  These savings result in an inference system able to outperform any of its numerical or neural component alone.  Experiments with stationary, smooth interfaces show that our hybrid framework is notably superior to conventional numerical methods in coarse grids and along steep interface regions.  Compared to our early developments in \cite{LALariosFGibou;LSCurvatureML;2021}, we have observed outstanding gains in precision after including training data pairs from more than a single interface type and transforming data with specialized input preprocessing.  In particular, our findings confirm that machine learning is a promising venue for reducing or removing mass loss in the level-set method.

We have organized this manuscript as follows.  Section \ref{sec:TheLevelSetMethod} summarizes the level-set method.  Then, Section \ref{sec:Methodology} gives a thorough description of our hybrid inference system and how to design enhanced neural networks for curvature computation.  We devote Section \ref{sec:ExperimentsAndResults} to experiments and results, and Section \ref{sec:Conclusions} draws some conclusions and opportunities for future work.


\colorsection{The level-set method}
\label{sec:TheLevelSetMethod}

The level-set method \cite{Osher1988} is an Eulerian formulation that defines an arbitrary interface of a domain, $\Omega \subseteq \mathbb{R}^n$, by $\Gamma \doteq \{\vv{x}: \phi(\vv{x}) = 0\}$ and its interior and exterior regions by $\Omega^- \doteq \{\vv{x}: \phi(\vv{x}) < 0\}$ and $\Omega^+ \doteq \{\vv{x}: \phi(\vv{x}) > 0\}$, where $\phi(\vv{x}): \mathbb{R}^n \mapsto \mathbb{R}$ is a Lipschitz continuous function known as the \emph{level-set function}.

Provided that $\phi(\vv{x})$ remains smooth after advection, it is straightforward to compute mean curvature, $\kappa(\vv{x})$, at any point $\vv{x} \in \Omega$ as

\begin{equation}
\kappa(\vv{x}) = \nabla \cdot \frac{\nabla \phi(\vv{x})}{|\nabla \phi(\vv{x})|},
\label{eq:numericalCurvature}
\end{equation}
where $\nabla \phi(\vv{x})$ is discretized with second-order accurate central differences.

In practice, $\phi(\vv{x})$ is periodically reinitialized into a signed normal distance function to the interface \cite{Sussman;Smereka;Osher:94:A-Level-Set-Approach} by solving the equation
\begin{equation}
\phi_\tau + S(\phi^0)(|\nabla\phi| - 1) = 0,
\label{eq:reinitialization}
\end{equation}
where $\tau$ represents pseudotime, $\phi^0$ is the level-set function before reshaping, and $S(\phi^0)$ is a smoothed-out signum function.  Reinitialization helps produce robust numerical results and simplifies computations \cite{Min:10:On-reinitializing-le}.  As shown in \cite{LALariosFGibou;LSCurvatureML;2021}, redistancing has also been linked to seamless neural network training and improved accuracy when estimating curvature along smooth interfaces.  For a complete description of the level-set method and its applications, we refer the interested reader to the classic text by Osher and Fedkiw \cite{Osher;Fedkiw:02:Level-Set-Methods-an} and the recent review by Gibou \etal \cite{GFO18}.


\colorsection{Methodology}
\label{sec:Methodology}

\begin{figure}[t]
	\centering
	\includegraphics[width=0.8\textwidth]{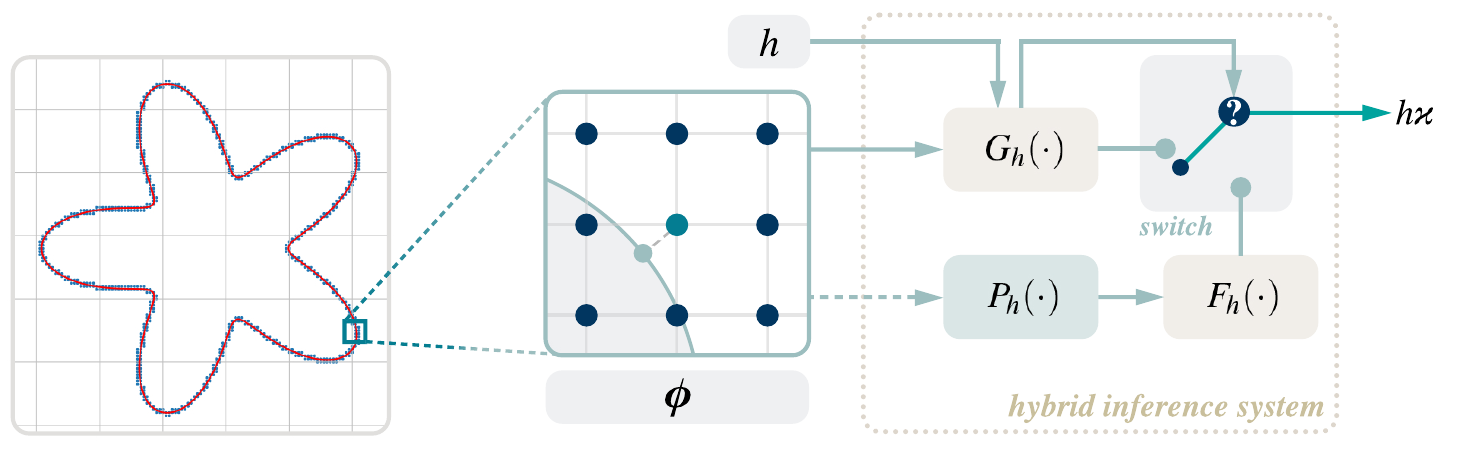}
	\caption{Overview of the hybrid inference system, its inputs, outputs, and functional elements: the compound numerical component, $G_h(\cdot)$, the preprocessing module, $P_h(\cdot)$, and the negative half curvature spectrum neural network, $F_h(\cdot)$.  Information flows through the solid arrows unrestrained, except at the dashed arrow, which is enabled on demand depending on the switching mechanism.  (Color online.)}
	\label{fig:blockDiagram}
\end{figure}

The present work refactors, extends, and improves the methodology of \cite{LALariosFGibou;LSCurvatureML;2021}, which applied deep learning to compute curvature in the level-set method.  Our results in \cite{LALariosFGibou;LSCurvatureML;2021} showed that, unlike machine learning and VOF methods \cite{CurvatureML19, VOFCurvature3DML19}, the level-set framework requires more complex supervised neural networks to approximate curvature values.  Also, training with samples collected from circular interfaces appears to yield less accurate estimations than a two-step numerical approach.  Here, we explore several ways to increase the curvature regression models' accuracy by training with distinct classes of two-dimensional interfaces and data augmentation \cite{A18, Hands-onMLwithScikit-LearnKerasAndTF19}.  In addition, we leverage machine learning techniques for input preprocessing \cite{scikit-learn11} and reduce the neural networks' spatiotemporal footprint by taking advantage of the problem's intrinsic symmetry.  These enhanced neural models are then selectively enabled in response to a switching mechanism that stands on conventional numerical schemes to gauge a first curvature approximation.  The block diagram shown in Figure \ref{fig:blockDiagram} illustrates our simple yet powerful hybrid inference system, including a numerical component, a preprocessing module, and a neural network.  Next, we describe each of these elements, devoting most of the details to the machine learning contribution.

\colorsubsection{Hybrid inference system}
\label{subsec:HybridInferenceSystem}

\begin{figure}[t]
	\centering
	\includegraphics[width=0.35\textwidth]{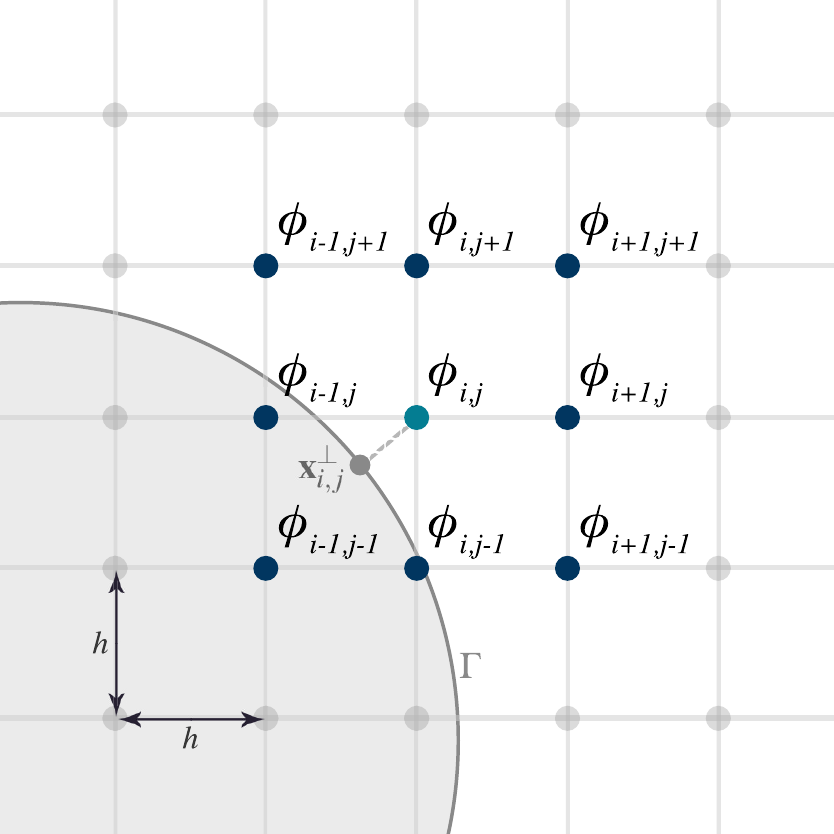}
	\caption{Nine-point stencil of node $(i,j)$ adjacent to the interface and the level-set function values involved in the numerical and neural estimation of curvature.  (Color online.)}
	\label{fig:stencil}
\end{figure}

An attractive advantage of numerical level-set methods is that they provide the simple relation \eqref{eq:numericalCurvature} for calculating $\kappa$ to any order of accuracy.  However, it is easy to see that the discretization of \eqref{eq:numericalCurvature} defines curvature everywhere in $\Omega$ but not precisely at $\Gamma$, where it is needed the most.  This problem has been linked to certain limitations when considering the well-balanced property of traditional level-set schemes \cite{Popinet;NumModelsOfSurfTension;18}.  Interpolation and distance-weighted average are standard formulations often used to bypass this deficiency and approximate curvatures closer to their expected values at the interface.  Therefore, we have adopted a \textit{compound numerical method}, denoted as the functional component $G_h(\cdot)$ in Figure \ref{fig:blockDiagram}, for estimating the \textit{dimensionless curvature}, $h\kappa$, at $\Gamma$.  $G_h(\cdot)$ consists of the second-order, central finite-difference discretization of \eqref{eq:numericalCurvature}, followed by bilinear interpolation at the location

\begin{equation}
\vv{x}_{i,j}^\Gamma = \vv{x}_{i,j} - \phi_{i,j}\frac{\nabla \phi(\vv{x}_{i,j})}{|\nabla \phi(\vv{x}_{i,j})|},
\label{eq:nodeAtInterface}
\end{equation}
where $\vv{x}_{i,j}^\Gamma$ is a rough approximation of $\vv{x}_{i,j}^\perp$, and $\phi_{i,j}$ is the level-set function value at node\footnote{We use the terms \textit{node}, \textit{vertex}, and \textit{grid point} as synonyms.} $(i,j)$.  Figure \ref{fig:stencil} contains a schematic of the $\phi$ values,

\begin{equation}
\vv{\phi} \doteq \begin{pmatrix}
	\phi_{i-1,j+1}, & \phi_{i,j+1}, & \phi_{i+1,j+1}, \\
	\phi_{i-1,j},   & \phi_{i,j},   & \phi_{i+1,j},   \\
	\phi_{i-1,j-1}, & \phi_{i,j-1}, & \phi_{i+1,j-1}
\end{pmatrix} \in \mathbb{R}^9,
\label{eq:stencil}
\end{equation}
involded in the discretization of \eqref{eq:numericalCurvature}.  More formally, $\vv{\phi}$ is an array of level-set values extracted from the standard nine-point stencil centered at the vertex $(i,j)$.  $h\kappa$ is the corresponding dimensionless curvature at the nearest point, $\vv{x}_{i,j}^\perp$, on the free boundary.

Our approach relies on crafting a dedicated hybrid inference system for each mesh resolution.  This process helps to reduce the complexity of the neural network in Figure \ref{fig:blockDiagram} by circumventing the intricacies involved in the construction of a universal model \cite{LALariosFGibou;LSCurvatureML;2021}.  Assuming that the computational domain, $\Omega$, has been discretized into the smallest local mesh size, $h = \Delta x = \Delta y$, we design and test a neural network, $F_h(\cdot)$, that fits exclusively to that nodal spacing.  This model is also suitable for regular and adaptive grids (see \cite{Strain1999, Min;Gibou:07:A-second-order-accur, Mirzadeh;etal:16:Parallel-level-set}).  We remark that we have resolved our computational domains with quad-trees \cite{BKOS00} for validating our hybrid inference system.  Such a discretization speeds up data set generation significantly when the level-set function used for training does not have an analytic form that one can evaluate on large grids (e.g., refer to Section \ref{subsec:SinusoidalInterfaceDataSetGeneration}).

\begin{figure}[t]
	\centering
	\includegraphics[width=0.65\textwidth]{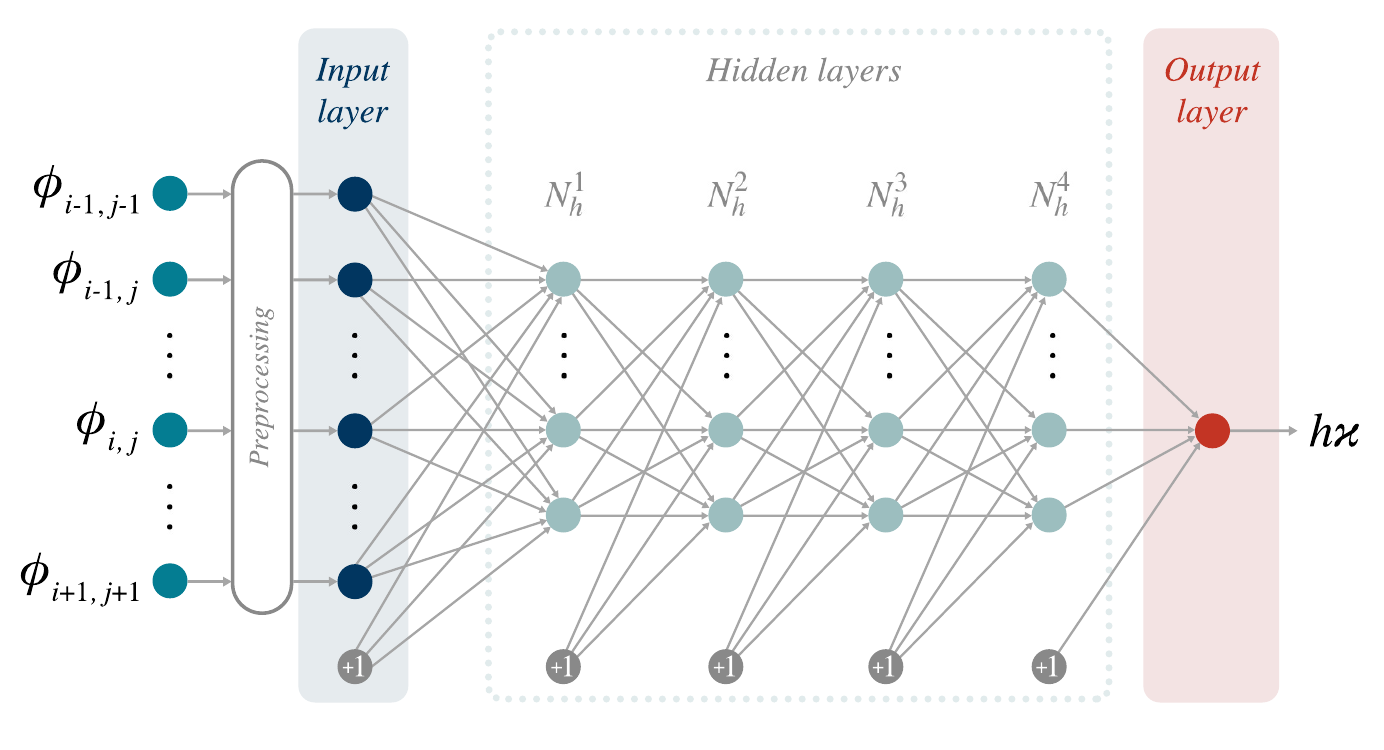}
	\caption{General architecture of our fully connected neural network, $F_h(\cdot)$, together with the preprocessing module, $P_h(\cdot)$.  The input to $F_h(\cdot)$ is a nine-dimensional array of preprocessed level-set function values.  Each hidden layer has $N_h^i$ ReLU units ($1 \leqslant i \leqslant 4$), and the output layer contains a single linear neuron that produces the dimensionless curvature, $h\kappa$, at the interface.  (Color online.)}
	\label{fig:mlp}
\end{figure}

Figure \ref{fig:mlp} outlines the architecture of $F_h(\cdot): \mathbb{R}^9 \mapsto \mathbb{R}$.  Our neural network is a supervised multilayer perceptron, trained on a synthetic data set, $\mathcal{D}$, that outputs the dimensionless curvature of $C^1$ interfaces in $\mathbb{R}^2$, given an input vector of (transformed) level-set values, $\vv{\phi}$.  Figure \ref{fig:mlp} also displays the preprocessing module, $P_h(\cdot)$, which transforms the inputs, $\vv{\phi}$, according to the patterns observed in $\mathcal{D}$ during training.  As we show in Section \ref{sec:ExperimentsAndResults}, $P_h(\cdot)$ plays a critical role in the hybrid inference system since it favors learning convergence and increases accuracy when we deal with coarse grids.

To assemble $\mathcal{D}$, we extract samples from \textit{sinusoidal-} and \textit{circular-interface} level-set functions.  By varying the configuration of such interfaces, one can construct a comprehensive collection of patterns that associate level-set function values to dimensionless curvatures.  In particular, a learning sample or data pair is a tuple $(\vv{\phi}, h\kappa)_{i,j}$, where $\vv{\phi}$ is given in \eqref{eq:stencil}, and $h\kappa$ is computed at $\vv{x}_{i,j}^\perp$ (see Figure \ref{fig:stencil}).  We collect these samples for any vertex $(i,j)$ that satisfies at least one of the four conditions: $\phi_{i,j} \cdot \phi_{i \pm 1,j} \leqslant 0$ or $\phi_{i,j} \cdot \phi_{i,j \pm 1} \leqslant 0$.  Sections \ref{subsec:SinusoidalInterfaceDataSetGeneration} and \ref{subsec:CircularInterfaceDataSetGeneration} below give a thorough description of the process for composing $\mathcal{D}$ from well-balanced, randomized data (sub)sets.

Next, we note that the curvature estimation problem features a subtle yet crucial property that we exploit when constructing $\mathcal{D}$ and designing $F_h(\cdot)$.  Namely, if one can estimate $h\kappa$ (with any solver) from the nine-point stencil $\vv{\phi}$, then $-\vv{\phi}$ should produce $-h\kappa$ under the same scheme.  It then suffices to store the negated stencils and their dimensionless curvatures in $\mathcal{D}$, incurring no information loss.  Following this idea, we build $\mathcal{D}$ from data pairs where the expected dimensionless curvature is always negative.  This technique is beneficial for three reasons: (1) it encourages spatiotemporal efficiency, (2) reduces topological complexity, and (3) improves the quality of the fitting process.  We thus hypothesize (and prove in Section \ref{subsec:FlowerShapeConvergenceStudy}) that training in such a \textit{negative half curvature spectrum} leads to an $F_h(\cdot)$ with the same or a higher expressive power than a model that handles negative and positive curvature samples.  This claim hinges on the availability of an inexpensive, reliable method to infer the curvature sign from the values in $\vv{\phi}$.  According to preliminary experimentation, $G_h(\cdot)$, the numerical approach we are trying to improve on, meets the requirements for a good sign indicator in all respects.

Now, let $F_h(\cdot)$ be a multilayer perceptron optimized for producing (negative) dimensionless curvatures in a computational domain with mesh size $h$.  Our proposed \textit{hybrid inference system} (Algorithm \ref{alg:hybridApproach}) is a decision-based process that couples the compound numerical method, $G_h(\cdot)$, a preprocessing module, $P_h(\cdot)$, and $F_h(\cdot)$ to estimate $h\kappa$ at the interface.  It uses $G_h(\cdot)$ as a first-level mechanism to gauge curvature and triggers $P_h(\cdot)$ and $F_h(\cdot)$ only if deemed necessary.  When the latter holds, the initial numerical approximation determines if the input stencil belongs in a convex or concave region.  Depending on this, the sign of the level-set values in \eqref{eq:stencil} is switched accordingly and passed onto $P_h(\cdot)$.  The multilayer perceptron then acts on the transformed input stencil and infers the signed $h\kappa$ at last.  As we prove in Section \ref{sec:ExperimentsAndResults}, the synergy of Algorithm \ref{alg:hybridApproach}'s elements is more effective in several aspects than using any of its components alone.

\begin{algorithm}[!t]
\small
\SetAlgoLined

\KwIn{mesh size, $h$; numerical component, $G_h(\cdot)$; preprocessing module, $P_h(\cdot)$; neural network, $F_h(\cdot)$; nine-point stencil of level-set values, $\vv{\phi}$.}
\KwResult{dimensionless curvature, $h\kappa$, at the interface, .}
\BlankLine

estimate $h\kappa$ using $G_h(\cdot)$\;

\eIf{$|h\kappa| < h\kappa_{flat}$}{ \label{alg:hybridApproach:condition}
	\Return $h\kappa$\;
}{
	use the numerical estimation of $h\kappa$ to determine its curvature sign, $\varsigma$\;

	\If{$\varsigma$ is $+$}{
		$\vv{\phi} \leftarrow -\vv{\phi}$\tcp*[r]{Flip sign of stencil of level-set values}
	}

	$\vv{\phi} \leftarrow P_h(\vv{\phi})$\;

	$h\kappa \leftarrow F_h(\vv{\phi})$\tcp*[r]{Neural inference}
	
	\If{$\varsigma$ is $+$}{
		$h\kappa \leftarrow -h\kappa$\tcp*[r]{Fix inference sign}
	}

	\Return $h\kappa$\;
}

\caption{Hybrid inference system for estimating dimensionless curvatures of two-dimensional, smooth interfaces.}
\label{alg:hybridApproach}
\end{algorithm}

To conclude this conceptual overview of the hybrid inference system, we briefly comment about the hyperparameter $\kappa_{flat}$ in Algorithm \ref{alg:hybridApproach}.  $\kappa_{flat}$ is a resolution-dependent, positive threshold that triggers the neural inference and helps discriminate flat regions where $G_h(\cdot)$ has proven to work well.  In that case, it is futile to call for a costly preprocessing and neural evaluation to improve numerical accuracy.  These savings are especially true if one has never exposed the multilayer perceptron to samples extracted from near planar or straight-line interfaces in $\mathcal{D}$.  Also, since we work on the negative half curvature spectrum, $\kappa_{flat}$ helps prevent producing unnecessary samples and reduce training complexity.  The reason behind this is that we can avoid generating or learning from data pairs where the target dimensionless curvature exceeds $-h\kappa_{flat}$.  However, it is always recommendable to give some room to the maximum expected output values in $\mathcal{D}$.  In other words, $\kappa \leqslant -C\kappa_{flat}$, $\forall(\vv{\phi}, h\kappa)$, and $0 < C < 1$, so that the neural network can smoothly interpolate curvatures near the upper bound, $-h\kappa_{flat}$.  In this work, we have chosen $\kappa_{flat} = 5$ and $C = \frac{1}{10}$.  To understand better how we use $\kappa_{flat}$ when assembling $\mathcal{D}$, we next describe the procedures for data set generation.  Then, Section \ref{subsec:TechnicalAspects} explains a few technical considerations for optimizing $F_h(\cdot)$ and preprocessing the samples in $\mathcal{D}$ with $P_h(\cdot)$.


\colorsubsection{Sinusoidal-interface data set generation}
\label{subsec:SinusoidalInterfaceDataSetGeneration}

Our novel approach comprises an enhanced training process that considers samples from level-set fields with sinusoidal zero isocontours.  The motivation for using such a class of elementary learning interfaces stems from the work of Despr\'{e}s and Jourdren \cite{DespresJourdren;MLDesignOfVOF;20}, which combined arcs, lines, and corners to design a machine-learning-flux procedure with satisfactory results.  In our case, sinusoids embody smooth compositions of all those patterns into malleable, periodic functions.  If needed, these parametrized sine waves can get as steep as possible to resemble a sequence of corners without losing smoothness.  Similarly, we have drawn inspiration from the stationary wave evaluated in \cite{CurvatureML19} and the authors' concluding remark about populating synthetic data sets with complex patterns extracted from simple generators.  Compared to \cite{LALariosFGibou;LSCurvatureML;2021}, in particular, sinusoids have helped improve accuracy considerably by spreading curvature values more evenly (and continuously) than the discrete distribution attained with circles.

\begin{figure}[t]
	\centering
	\begin{subfigure}[b]{0.4\textwidth}
		\includegraphics[width=\textwidth]{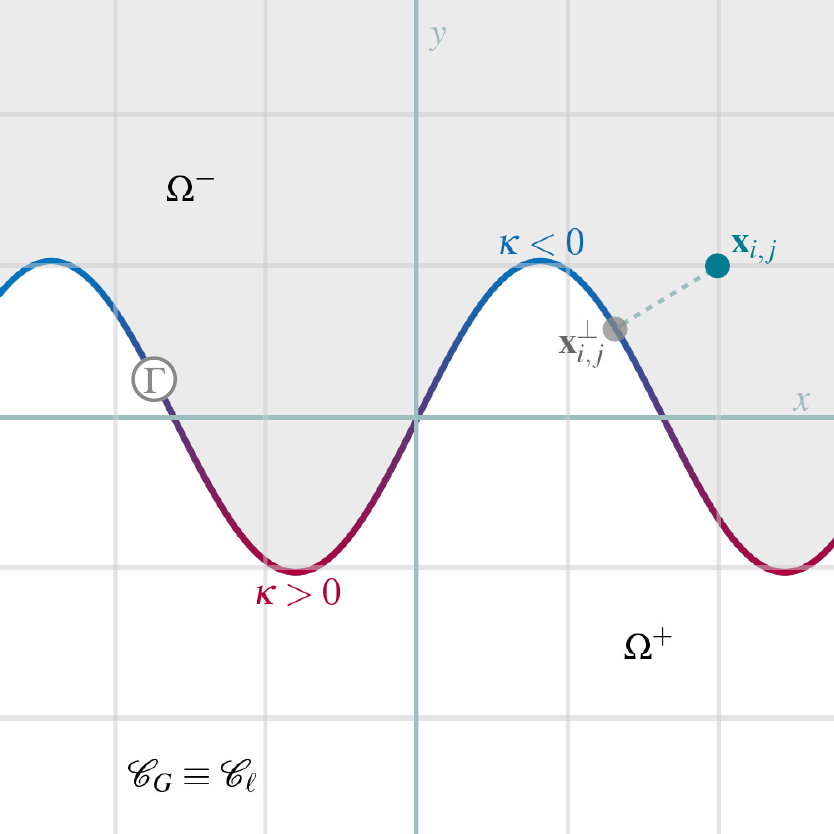}
		\caption{\footnotesize Canonical $f_s(t) = A\sin{(\omega t)}$}
		\label{fig:sine1}
	\end{subfigure}
	~
	\begin{subfigure}[b]{0.4\textwidth}
		\includegraphics[width=\textwidth]{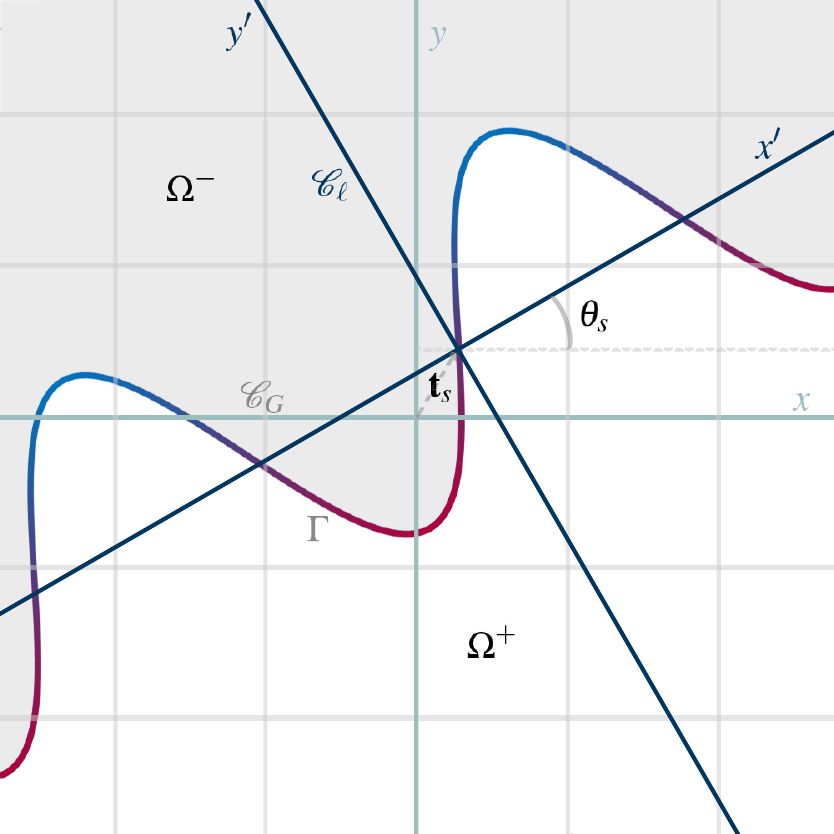}
		\caption{\footnotesize Affine transformation of sinusoidal interface}
		\label{fig:sine2}
	\end{subfigure}
	\caption{The sinusoidal interface, $f_s(t)$, its local coordinate system, $\mathcal{C}_\ell$, and the partitioning of the computational domain into negative ($\Omega^-$) and positive ($\Omega^+)$ regions.  (Color online.)}
	\label{fig:sineInterface}
\end{figure}

Let $\mathcal{D}_s$ be the set of learning data pairs, $(\vv{\phi}, h\kappa)$, collected from bivariate functions, $\phi(\vv{x}): \mathbb{R}^2 \mapsto \mathbb{R}$, whose zero level sets are expressed as

\begin{equation}
f_s(t) = A\sin{(\omega t)},
\label{eq:sineInterface}
\end{equation}
where $A$ is the wave amplitude, and $\omega$ is the sinusoidal frequency.  To construct $\phi(x,y)$, we start by parametrizing and discretizing $f_s(t)$ into line segments.  This strategy is similar to how animators control the motion of points along curves \cite{ComputerAnimation08}.  After discretization is complete, we define the level-set function

\begin{equation}
\phi_{rls}^s(x, y) = 
	\begin{cases}
	-d_s(x, y) & \textrm{if } y > f_s(x) \\
	+d_s(x, y) & \textrm{if } y < f_s(x) \\
	 0 & \textrm{if } y = f_s(x)
	\end{cases},
\label{eq:sineLevelSet}
\end{equation}
where $d_s(x, y)$ is the shortest distance between $(x, y)$ and the closest line segment on the interface.  Notice that $\Gamma$ splits the computational domain, $\Omega$, into negative and positive regions depending on whether a vertex lies above or below $f_s(x)$ (see Figure \ref{fig:sine1}).  We adopt this configuration because it ensures that curvature comes up positive along convex portions of the negative domain.  This choice also agrees with the analytic expression that yields the sine wave's signed curvature,

\begin{equation}
\kappa_s(t) = -\frac{A\omega^2 \sin{(\omega t)}}{\left(1 + A^2 \omega^2 \cos^2{(\omega t)}\right)^{3/2}}.
\label{eq:sineCurvature}
\end{equation}

Given a node with coordinates $\vv{x}_{i,j} = (x_i, y_i)$ next to the interface, we apply bisection and Newton--Raphson's root finding to recover the optimal $t^*$ that produces $\vv{x}_{i,j}^\perp$ in Figures \ref{fig:stencil} and \ref{fig:sine1}.  Once $t^*$ is available, we compute the target curvature with \eqref{eq:sineCurvature}.  To ensure that $\phi_{rls}^s(x, y)$ resembles a signed distance function, we reinitialize it by solving \eqref{eq:reinitialization} to steady state.  This way, \eqref{eq:sineLevelSet} is assured to turn out well regularized as long as one discretizes $f_s(t)$ with sufficiently small parameter intervals, $\Delta t < h$.  We have chosen ten iterations throughout this work to solve \eqref{eq:reinitialization} numerically.  This choice has resulted in a cheap redistancing process that guarantees enough smoothness in the level-set function.  As pointed out in \cite{LALariosFGibou;LSCurvatureML;2021}, reinitialization is fundamental for removing unstructured noise that worsens architectural complexity and undermines accuracy. Also, neural networks improve their interpolative power when one includes redistanced values and exact signed distances to $\Gamma$.  Such an improvement is possible because the regression model can better assimilate connections between noiseless data pairs and their reinitialized versions when both share the same $h\kappa$ output.

In light of the previous observation, besides sampling $\phi_{rls}^s(x,y)$ over $\Omega$, we extract data pairs from an exact signed distance function, $\phi_{sdf}^s(x, y)$.  Specifying the latter relies entirely on numerical optimization for computing the shortest distance from grid points to $\Gamma$.  Then again, for each node $(i, j)$ next to the interface, the optimal parameter $t^*$ that determines $\vv{x}_{i,j}^\perp$ is used to compute $\kappa_s(t)$ with \eqref{eq:sineCurvature}.  This process allows us to assemble signed distance samples mirroring their reinitialized versions to make up $\mathcal{D}_s$.

\textbf{Remark}:  An alternative method to generate reinitialized sinusoidal-interface samples involves ``redistancing'' the signed distance function $\phi_{sdf}^s(x, y)$.  This process works because reinitialization introduces numerical perturbations that our algorithm can interpret as noisy inputs.  Depending on the mesh size, it is often faster to produce reinitialized data this way than parametrizing $\Gamma$ to find distances iteratively. 

One can manipulate the shape parameters $A$ and $\omega$ in \eqref{eq:sineInterface} to generate a wide variety of patterns involving $\vv{\phi}$ and $h\kappa$.  However, greater diversification is possible if, besides amplitude and frequency, we also perturb the local frame of the sinusoidal interface.  Figure \ref{fig:sine2} depicts this class of data augmentation, which comprises the translation and rotation of the interface coordinate system, $\mathcal{C}_\ell$.  To do this, we could redefine the level-set functions $\phi_{sdf}^s(x, y)$ and $\phi_{rls}^s(x, y)$ in a way that they would take a translation vector, $\vv{t}_s = \left(x_0, y_0\right)$, and a rotation angle, $\theta_s$, as additional inputs.  In practice, estimating distances from points to the simpler, canonical function $f_s(t)$ remains unaltered even when $\vv{t}_s$ and $\theta_s$ are present.  To illustrate this idea, let $\mathcal{C}_G$ be the global coordinate system, and let $T(\cdot,\cdot)$ and $R(\cdot)$ be $3 \times 3$ translation and rotation matrices in homogeneous coordinates \cite{CGUsingOpenGL01, ComputerAnimation08}.  We claim that if $_G{M}_\ell = R(\theta_s)T(x_0, y_0)$ is an affine transformation that maps points from the local to the global coordinate system, then $_\ell{M}_G = {}_G{M}_\ell^{-1} = T(-x_0, -y_0)R^T(\theta_s)$ maps points in the reverse direction: from $\mathcal{C}_G$ to $\mathcal{C}_\ell$.  Suppose that we express the point $\vv{x}_G = (x_G, y_G)^T$ in terms of $\mathcal{C}_G$, then

\begin{equation}
\vv{x}_\ell = 
\begin{pmatrix}
	x_\ell \\
	y_\ell 
\end{pmatrix} = 
\begin{pmatrix}
	\cos\theta_s(x_G - x_0) + \sin\theta_s(y_G - y_0) \\
	\cos\theta_s(y_G - y_0) + \sin\theta_s(x_0 - x_G)
\end{pmatrix}
\label{eq:GlobalToLocalTransformation}
\end{equation}
are the coordinates of $\vv{x}_G$ with respect to the local coordinate system, $\mathcal{C}_\ell$.  Having $\vv{x}_\ell = (x_\ell, y_\ell)^T$ at hand, one can evaluate $\phi_{sdf}^s$ and $\phi_{rls}^s$ as usual to find $t^*$ and compute the exact distance and curvature at the closest point on $\Gamma$.

\begin{algorithm}[!t]
\small
\SetAlgoLined
\SetKwFunction{linspace}{LinSpace}
\SetKwFunction{levelset}{LevelSet}
\SetKwFunction{levelsetSDF}{LevelSetSDF}
\SetKwFunction{randomsamples}{RandomSamples}
\SetKwFunction{easein}{EaseIn}

\KwIn{mesh size, $h$; minimum curvature, $\kappa_{min}$; maximum curvature, $\kappa_{max}$.}
\KwResult{data set with samples from sinusoidal interfaces, $\mathcal{D}_s$.}
\BlankLine
$h_{base} \leftarrow 2^{-7}$\tcp*[r]{Reference mesh size}
$L_p = 1 + \frac{\log_2( h_{base} / h )}{3}$\tcp*[r]{Linear proportion as $h$ varies from $h_{base}$ to $2^{-10}$}
\BlankLine

$N_a \leftarrow 33$\tcp*[r]{Number of distinct amplitudes}
$A_{min} \leftarrow 1.5h, \quad A_{max} \leftarrow 0.25$\tcp*[r]{Amplitude bounds}
\BlankLine

$N_\theta \leftarrow 34$\tcp*[r]{Number of distinct tilts of sine's main axis}
$\theta_{min} \leftarrow -\pi/4, \quad \theta_{max} \leftarrow \pi/4$\tcp*[r]{Tilt bounds}
\BlankLine

$\mathcal{D}_s \leftarrow \emptyset$\;
\BlankLine

\tcp{Evaluate $N_a$ equally spaced sine amplitudes $A \in [A_{min}, A_{max}]$}
\ForEach{amplitude $A$ in \linspace{$A_{min}$, $A_{max}$, $N_a$}}{
	\tcp{Vary $\max{(\kappa)}$ for current amplitude $A$ between $\kappa_{max}/4$ and $\kappa_{max}$}
	$\omega_{max} \leftarrow \sqrt{\frac{\kappa_{max}}{A}}, \quad \omega_{min} \leftarrow \sqrt{\frac{\kappa_{max}}{4A}}$\tcp*[r]{Frequency range}
	$\mu_{\max{(\kappa)}} \leftarrow \frac{1}{2}\left(\frac{\kappa_{max}}{4} + \kappa_{max}\right)$\tcp*[r]{Midpoint between $\max{(\kappa)}$ lower and upper bounds}
	$d_{crests} \leftarrow \frac{\pi}{2}\left( \frac{1}{\omega_{min}} - \frac{1}{\omega_{max}} \right)$\tcp*[r]{$t$-distance between first crests at $\omega_{min}$ and $\omega_{max}$}
	$N_\omega \leftarrow \left\lceil \frac{d_{crests}}{h_{base}}L_p \right\rceil + 1$\tcp*[r]{Number of distinct frequency values}
	\BlankLine
	
	\tcp{Evaluate $N_\omega$ equally spaced sine frequencies $\omega \in [\omega_{min}, \omega_{max}]$}
	\ForEach{frequency $\omega$ in \linspace{$\omega_{min}$, $\omega_{max}$, $N_\omega$}}{
		\tcp{Data sets for exact signed distance function}
		\tcp{and reinitialized level-set function}
		$\mathcal{S}_{sdf} \leftarrow \emptyset, \quad \mathcal{S}_{rls} \leftarrow \emptyset$\;
		\BlankLine
		
		\tcp{Evaluate $N_\theta - 1$ equally spaced sine tilts $\theta_s \in [\theta_{min}, \theta_{max})$}
		\ForEach{tilt $\theta_s$ in \linspace{$\theta_{min}$, $\theta_{max}$, $N_\theta$}, except $\theta_{max}$}{
			$(x_0, y_0) \sim \mathcal{U}(-h/2, +h/2)$\tcp*[r]{Random translation of $\mathcal{C}_\ell$}
			$\phi_{sdf}^s \leftarrow$ \levelsetSDF{$A$, $\omega$, $\theta_s$, $x_0$, $y_0$}\tcp*[r]{Build level-set functions}
			$\phi_{rls}^s \leftarrow$ \levelset{$A$, $\omega$, $\theta_s$, $x_0$, $y_0$}\;
			\BlankLine
			
			set up computational domain, $\Omega \equiv [-0.5, 0.5]^2$\;

			evaluate $\phi_{sdf}^s$ and $\phi_{rls}^s$ on discretized $\Omega$\;

			reinitialize $\phi_{rls}^s$ using 10 iterations\;
			\BlankLine
			
			\ForEach{node $n$ along $\Gamma$}{
				compute $n$'s target $\kappa$ using the transformed sinusoidal interface $f_s(t) = A\sin{(\omega t)}$\;

				\lIf{$|\kappa| < \kappa_{min}$}{
					skip $n$ as $\kappa$ belongs to a flat region
				}
				\BlankLine
				
				\tcp{Take samples from $n$ with an ease-in probability from 0.05 to 1}
				\tcp{where $Pr(|\kappa| >= \mu_{\max{(\kappa)}}) = 1$ and $Pr(|\kappa| = 0) = 0.05$}
				\If{$\mathcal{U}(0, 1) < 0.05 + 0.95*$ \easein{$|\kappa| / \mu_{\max{(\kappa)}}$}}{
					build pairs $(-1)^{\kappa/|\kappa|}(\vv{\phi}_{sdf}, h\kappa)$ and $(-1)^{\kappa/|\kappa|}(\vv{\phi}_{rls}, h\kappa)$ from $n$'s stencil\;

					augment samples by rotating stencils by $90^\circ$, three times\;

					add augmented samples to $\mathcal{S}_{sdf}$ and $\mathcal{S}_{rls}$\;
				}
			}
		}
		
		$\mathcal{D}_s \leftarrow \mathcal{D}_s \;\cup\; \mathcal{S}_{sdf} \;\cup\; \mathcal{S}_{rls}$\tcp*[r]{Accumulate samples in output set}
	}
}
\BlankLine
	
\Return $\mathcal{D}_s$\;

\caption{Data set generation from sinusoidal-interface level-set functions.}
\label{alg:samplingSines}
\end{algorithm}
  
The procedure for assembling $\mathcal{D}_s$ is combinatorial, with some randomization and post-processing steps that fight data explosion.  The methodology also incorporates some heuristics to construct a balanced data set.  \textit{Balancing data sets} is a staple concept in pattern recognition applications.  The idea behind this is to provide each target class with the same probability of being accounted for by the learning process.  In our case, these target classes correspond to the continuous curvature values associated with $\vv{\phi}$ stencils.  Algorithm \ref{alg:samplingSines} summarizes the strategy for collecting samples from sinusoidal-interface level-set functions.  The parameters one can vary to produce different boundary configurations are amplitude, $A$, frequency, $\omega$, displacement, $\vv{t}_s$, and tilt, $\theta_s$.  These depend on the minimum and maximum target curvatures, $\kappa_{min} = \frac{1}{2}$ and $\kappa_{max} = 85\frac{1}{3}$, and the computational mesh size, $h$.  Recalling what we stated at the end of Section \ref{subsec:HybridInferenceSystem}, the value for $\kappa_{min}$ comes from setting $\kappa_{flat} = 5$ and $C = \frac{1}{10}$.  Furthermore, as we point out below, one must be very careful with sinusoidal interfaces at all times because flat regions dominate every single parameter choice.

At the outermost layer, we evaluate $N_a$ equally spaced amplitudes, $A$, ranging from $1.5h$ to $0.25$.  In the following layer, we define the sinusoidal frequency, $\omega$, which varies between $\omega_{min}$ and $\omega_{max}$.  This interval guarantees that the maximum (absolute) curvature along the crests lies within the range of $\left[\frac{1}{4}\kappa_{max}, \kappa_{max}\right]$.  The number of discrete frequency steps, $N_\omega$, is proportional to the $t$-distance between the first positive crests at frequencies $\omega_{min}$ and $\omega_{max}$ and inversely proportional to a reference mesh size, $h_{base}$.  This base mesh size helps to dampen the overproduction of samples as $h \rightarrow 0$.  Further, $N_\omega$ is scaled by a constant, $L_p$, which increases linearly with $h$ to account for smaller local cell widths.

After choosing $A$ and $\omega$, we perturb the local reference frame, $\mathcal{C}_\ell$, that embeds the canonical sinusoidal interface \eqref{eq:sineInterface}.  This task begins by varying the tilt, $\theta_s$, over the range of $[-\pi/4, +\pi/4)$, with $N_\theta$ uniform values.  Next, the displacement, $\vv{t}_s = (x_0, y_0)$, is extracted from a uniform random distribution whose lower and upper bounds are $-h/2$ and $+h/2$.  Having thus defined all the interface shape and transformation parameters, we prepare $\phi_{rls}^s(x,y)$ and $\phi_{sdf}^s(x,y)$ and sample them on the discretized computational domain.  As noted above, the data pairs in $\mathcal{D}_s$ come from exact signed distance functions and their reinitialized versions.

The following step in Algorithm \ref{alg:samplingSines} involves analyzing nodes along $\Gamma$ and deciding whether we should add their data pairs to $\mathcal{D}_s$ or discard them.  We seek to prioritize steep curvatures over flat ones to build a balanced data set as much as possible.  First, we filter out any vertex $(i, j)$ if its projection, $\vv{x}_{i,j}^\perp$, belongs in a sufficiently planar region ($|\kappa| < \kappa_{min}$).  If the target $|\kappa|$ at the boundary is above the threshold $\mu_{\max{(\kappa)}} = \textrm{mean}\left(\frac{1}{4}\kappa_{max}, \kappa_{max}\right)$, we keep the grid point.  For the remaining candidate vertices, we use an ease-in \cite{ComputerAnimation08} cumulative distribution function to determine if they should be part of $\mathcal{D}_s$.  This probability depends on the ratio $|\kappa| / \mu_{\max{(\kappa)}}$ and ensures that high curvatures are more likely to make it to the final step.  Next, we assemble the pairs $(-1)^{\kappa/|\kappa|}(\vv{\phi}, h\kappa)$ for the selected vertices, restating that we always store the negative curvature version of any sample.  Lastly, we resort to extra data augmentation by rotating each stencil by $90^\circ$, up to three times.  This technique is not uncommon when training image classifiers \cite{Hands-onMLwithScikit-LearnKerasAndTF19}, and we have introduced it in our strategy to enhance model generalization.

Experience has shown that despite the probabilistic approach described above, the $\kappa$ distribution in $\mathcal{D}_s$ tends to get skewed near $-\kappa_{min}$.  To mitigate this problem, we perform a second round of random selection with \textit{sample binning}.  The procedure is simple.  First, the elements in $\mathcal{D}_s$ get organized into a histogram with 20 or 40 containers.  The grouping criterion is the dimensionless curvature.  Next, one finds the size, $N_b$, of the bin with the smallest number of data pairs.  Then, we sub-sample any bin of size larger than $K\cdot N_b$, for $K = 2, 3.5$, (with no replacement) until the number of points in it does not exceed $K\cdot N_b$.  This makes sure that the target curvature histogram of $\mathcal{D}_s$ looks more or less rectangular.


\colorsubsection{Circular-interface data set generation}
\label{subsec:CircularInterfaceDataSetGeneration}

In addition to training on sinusoidal-interface samples, we also consider data pairs $(\vv{\phi}, h\kappa)$ extracted from level-set functions with circular zero isocontours.  The procedure for collecting these learning samples resembles the method outlined in \cite{CurvatureML19} and \cite{LALariosFGibou;LSCurvatureML;2021}.  Here, however, we introduce a distinct approach for spacing out circles across the domain and ensure that all discrete curvatures are sampled (more or less) equally.  As alluded to in the preceding Section \ref{subsec:SinusoidalInterfaceDataSetGeneration}, the goal is to give all curvature values the same probability of being considered by the learning algorithm.  To achieve this, we do not collect samples from a fixed number of circular interfaces with the same radius \cite{LALariosFGibou;LSCurvatureML;2021}.  Instead, we explicitly set the number of data pairs that we must extract for every discrete radius.  This technique increases the visibility of high-curvature data pairs (i.e., from small-radius circles); had we not done that, such samples would be under-represented given that low-curvature patterns can quickly become dominant.

Let $\mathcal{D}_c$ be the set of data pairs extracted from bivariate functions, $\phi(\vv{x}) : \mathbb{R}^2 \mapsto \mathbb{R}$, with circular isocontours.  Define the level-set functions

\begin{equation}
\phi_{sdf}^c(x, y) = \sqrt{\left(x - x_0\right)^2 + \left(y - y_0\right)^2} - r 
\quad \textrm{and} \quad
\phi_{rls}^c(x, y) = \left(x - x_0\right)^2 + \left(y - y_0\right)^2 - r^2, 
\label{eq:circularInterfaces}
\end{equation}
where $(x_0, y_0)$ is the center of the circular interface, and $r$ is the radius.  Notice that unlike $\phi_{sdf}^c(x, y)$, $\phi_{rls}^c(x, y)$ is not a signed distance function.  Thus, one must reinitialize $\phi_{rls}^c(x, y)$ after evaluating the level-set function on the discretized domain.

Algorithm \ref{alg:samplingCircles} describes how to construct $\mathcal{D}_c$.  Once more, we feed the process with the mesh size and the minimum and maximum curvature values of $\kappa_{min} = \frac{1}{2}$ and $\kappa_{max} = 85\frac{1}{3}$.  After setting these constants, generating the data pairs $(\vv{\phi}, h\kappa)$ boils down to deciding how many to collect along $\Gamma$ and how to distribute the interfaces across $\Omega$.  

\begin{algorithm}[!t]
\small
\SetAlgoLined
\SetKwFunction{linspace}{LinSpace}
\SetKwFunction{levelset}{LevelSet}
\SetKwFunction{levelsetSDF}{LevelSetSDF}
\SetKwFunction{randomsamples}{RandomSamples}

\KwIn{mesh size, $h$; minimum curvature, $\kappa_{min}$; maximum curvature, $\kappa_{max}$.}
\KwResult{data set with samples from circular interfaces, $\mathcal{D}_c$.}
\BlankLine

$h_{base} \leftarrow 2^{-7}$\tcp*[r]{Reference mesh size}

$r_{min} \leftarrow 1/\kappa_{max}$\tcp*[r]{Minimum and maximum radii}

$r_{max} \leftarrow 1/\kappa_{min}$\;

$\kappa_{flat} \leftarrow 10\kappa_{min}$ \tcp*[r]{Minimum $\kappa$ and radius to trigger neural inference}

$r_{flat} \leftarrow 1/\kappa_{flat}$\;

\BlankLine

$N_r \leftarrow \left\lceil 2\left( \frac{r_{max} - r_{min}}{h_{base}} + 1 \right)\left(\log_2\left( \frac{h_{base}}{h} \right) + 1\right) \right\rceil$\tcp*[r]{Number of distinct radii}
$N_s \leftarrow \left\lceil \frac{5\pi}{h^2} \left(r_{flat}^2 - \left(r_{flat} - h\right)^2\right) \right\rceil$\tcp*[r]{Potential number of samples per radius}
$\bar{N}_s \leftarrow \left\lceil \frac{5\pi}{h_{base}^2} \left(r_{flat}^2 - \left(r_{flat} - h_{base}\right)^2\right) \right\rceil$\tcp*[r]{Allowable samples per radius}

$\mathcal{D}_c \leftarrow \emptyset$\;
\BlankLine

\tcp{Evaluate $N_r$ radii computed from equally spaced $\kappa \in [\kappa_{min}, \kappa_{max}]$}
\ForEach{curvature $\kappa$ in \linspace{$\kappa_{min}$, $\kappa_{max}$, $N_r$}}{
	$r \leftarrow 1/\kappa$\tcp*[r]{Current radius}
	\BlankLine

	\tcp{Data sets for exact signed distance function}
	\tcp{and reinitialized level-set function}
	$\mathcal{S}_{sdf} \leftarrow \emptyset$, $\mathcal{S}_{rls} \leftarrow \emptyset$\;	
	\BlankLine
	
  	\While(\tcp*[f]{Collect samples for current radius}){$|\mathcal{S}_{sdf}| < N_s$}{
		$(x_0, y_0) \sim \mathcal{U}(-h/2, +h/2)$\tcp*[r]{Random circle center around origin}
		$\phi_{sdf}^c \leftarrow$ \levelsetSDF{$x_0$, $y_0$, $r$}\tcp*[r]{Build level-set functions}
		$\phi_{rls}^c \leftarrow$ \levelset{$x_0$, $y_0$, $r$}\;
		\BlankLine
		
		set up computational domain, $\Omega$, with proper bounds\;

		evaluate $\phi_{sdf}^c$ and $\phi_{rls}^c$ on discretization of $\Omega$\;

		reinitialize $\phi_{rls}^c$ using 10 iterations\;
		\BlankLine
		
		\ForEach{node $n$ along $\Gamma$}{
			build pairs $-(\vv{\phi}_{sdf}, h\kappa)$ and $-(\vv{\phi}_{rls}, h\kappa)$ from the stencil of $n$\;

			add samples to $\mathcal{S}_{sdf}$ and $\mathcal{S}_{rls}$
		}
	}
	\BlankLine

	\tcp{Extract $\bar{N}_s$ samples from $\mathcal{S}_{sdf}$ and $\mathcal{S}_{rls}$ and add them to $\mathcal{D}_c$}   
	$\mathcal{D}_c \leftarrow \mathcal{D}_c \;\cup$ \randomsamples{$\mathcal{S}_{sdf}$, $\bar{N}_s$}\;

	$\mathcal{D}_c \leftarrow \mathcal{D}_c \;\cup$ \randomsamples{$\mathcal{S}_{rls}$, $\bar{N}_s$}\;
}
\BlankLine
	
\Return $\mathcal{D}_c$\;

\caption{Data set generation from circular-interface level-set functions.}
\label{alg:samplingCircles}
\end{algorithm}

We begin by specifying the number of unique radii, $N_r$.  This is necessary to define the level-set functions in \eqref{eq:circularInterfaces}. Intuitively, $N_r$ should be proportional to the mesh size $h$; however, such a number could become prohibitive as $h \rightarrow 0$.  We circumvent this issue by expressing $N_r$ in terms of the base resolution, $h_{base}$, and requiring a linear growth rather than exponential.  As for the number of samples, $N_s$, to collect for each interface radius, we approximate it with the area difference between the circle with radius $\kappa_{flat}^{-1}$ and the following circle whose radius is $ (\kappa_{flat} - h)^{-1}$.  Notice we have opted for a radius smaller than $\kappa_{min}^{-1}$ to avoid overproducing data pairs.  Afterward, we only keep a random subset of size $\bar{N}_s$ to further bound the training population.  This whole system guarantees that, at least for $\mathcal{D}_c$, each discrete $\kappa$ value is equally probable of being observed during training.

Next, one must resolve how to space out $N_r$ unique radii across $\Omega$.  Unlike \cite{CurvatureML19} and \cite{LALariosFGibou;LSCurvatureML;2021}, we compute such radii from uniform curvature values between $\kappa_{min}$ and $\kappa_{max}$.  The rationale for this choice abides by the balanced data set principle; had we chosen to calculate $\kappa$ from equidistant radii, we would have gotten the target dimensionless curvature distribution in $\mathcal{D}_c$ overly skewed near zero. 

Algorithm \ref{alg:samplingCircles} also shows how to introduce randomness into $\mathcal{D}_c$ when defining the level-set functions $\phi_{sdf}^c(x, y)$ and $\phi_{rls}^c(x, y)$.  To do this, we perturb the circles' centers about the origin of $\Omega$ with a uniform random value between $-h/2$ and $+h/2$.  Notice that the latter matches the perturbation method used to affine-transform sinusoidal interfaces in Section \ref{subsec:SinusoidalInterfaceDataSetGeneration}.  This simple form of data augmentation has proven effective in enhancing the generalization of other inference systems, such as image classifiers and convolutional networks \cite{A18, Hands-onMLwithScikit-LearnKerasAndTF19}.  In our approach, we repeatedly generate random interface centers until the expected number of samples is satisfied.

After the sinusoidal- and circular-interface data sets are complete, we combine them into $\mathcal{D} \leftarrow \mathcal{D}_c \cup \mathcal{D}_s$ and continue to train a dedicated neural network, $F_h(\cdot)$, for the selected mesh size.


\colorsubsection{Technical aspects}
\label{subsec:TechnicalAspects}

Training $F_h(\cdot)$ is algorithmically simpler but much more resource-intensive than any of the tasks described thus far.  Given a cell width, $h$, we design and train the corresponding multilayer perceptron on the negative-curvature data set, $\mathcal{D}$, using TensorFlow \cite{Tensorflow15} and Keras \cite{Keras15}, in Python.  We follow customary practices \cite{Mehta19, A18} and split $\mathcal{D}$ into training, testing, and validation subsets using 70\%, 15\%, and 15\% of the data pairs, respectively.  As a technical note, we generate all the samples collected in Algorithms \ref{alg:samplingSines} and \ref{alg:samplingCircles} with our in-house C++ implementation of the parallel adaptive level-set method of Mirzadeh \etal \cite{Mirzadeh;etal:16:Parallel-level-set}.  This implementation also contains tools to realize the numerical component, $G_h(\cdot)$, and perform accuracy assessments for any interface.

As seen in Figure \ref{fig:mlp}, the inputs to $F_h(\cdot)$ are the preprocessed versions of the level-set values in $\vv{\phi}$.  Preprocessing is not only crucial during training but also during testing and deployment.  The preprocessing module, $P_h(\cdot)$, is based on the training subset and is used to transform raw feature vectors into a more suitable representation for the regression model \cite{scikit-learn11}.  In particular, as shown by LeCun \etal in \cite{LeCun;EfficientBackProp;98}, we have discovered that centering, scaling, and uncorrelating features in $\vv{\phi}$ via \textit{principal component analysis} (PCA) and \textit{whitening} speed up convergence and delivers better results than typical standardization (see \cite{LALariosFGibou;LSCurvatureML;2021}).  

A PCA transformation involves a change of coordinates, where the new basis vectors are the principal components of the $n\times 9$ training subset $D \subset \mathcal{D}$ \cite{Parker;CS170A;2016}, and $n$ is the number of training samples (see \eqref{eq:stencil}).  \emph{Before network-weight optimization}, we find these principal components by initially computing $D$'s covariance matrix $C \in \mathbb{R}^{9 \times 9}$.  More succinctly, $C = \frac{1}{n-1}(D-M)^T(D-M)$, where $M$ is an $n \times 9$ matrix containing $D$'s column-wise mean vector $\vv{\mu} \in \mathbb{R}^9$ stacked $n$ times.  With $C$ at hand, we then retrieve its SVD decomposition $C = USV^T$, which finally uncovers $D$'s principal components represented by the columns in $V$.  Later, to transform data, we first use $V$ to recast any centered vector $\vv{\varphi} = \vv{\phi} - \vv{\mu}$ into $\vv{\phi}'$ by simply projecting $\vv{\varphi}$ onto the new basis.  In a second step, we incorporate whitening by point-wise dividing $\vv{\phi}'$ by $\sqrt{\vv{\sigma}}$, where $\vv{\sigma} \in \mathbb{R}^9$ holds the singular values (i.e., variances) in $S$.  The idea behind whitening is to rescale uncorrelated features to the same level of importance so that the neural estimator can decide which of them to emphasize more easily \cite{A18}.  Algorithm \ref{alg:Preprocessing} summarizes these transforming steps embodied in $P_h(\cdot)$, which we apply to the learning (i.e., training, testing, and validation) subsets and \emph{at inference time}.  In our case, we have used SciKit-Learn's \texttt{PCA} class \cite{scikit-learn11} to extract the principal components and carry out data transformation.  Also, python's pickle object serialization has helped us port the $P_h(\cdot)$ module across applications.


\begin{algorithm}[!t]
\small
\SetAlgoLined
\SetKwFunction{getstats}{GetStats}

\KwIn{negative-curvature-normalized nine-point stencil of level-set values, $\vv{\phi}$.}
\KwResult{transformed feature vector, $\vv{\phi}'$.}
\BlankLine

\tcp{In-place PCA transformation and whitening}
$Q \leftarrow$ \getstats{}\tcp*[r]{Retrieve training pickle objects}
let $V$, $\vv{\sigma}$, and $\vv{\mu}$ be the components' matrix, the standard deviations' vector, and the feature means' vector in $Q.${\tt PCA}\;
$\vv{\phi}' \leftarrow (V^T(\vv{\phi}-\vv{\mu})) \oslash \vv{\sigma}$\tcp*[r]{PCA transformation and whitening ($\oslash$ is an element-wise division)}
\BlankLine

\Return $\vv{\phi}'$\;

\caption{Preprocessing module $P_h(\cdot)$.}
\label{alg:Preprocessing}
\end{algorithm}

To conclude this section, we note that all of our networks for solving the curvature problem exhibit the five-layered architecture already shown in Figure \ref{fig:mlp}.  The input and output neurons of $F_h(\cdot)$ are linear, and all the computing nodes at the intermediate, hidden layers are nonlinear ReLU units.  Thus, a forward pass on $F_h(\cdot)$, given an input vector $\vv{\phi} \in \mathbb{R}^9$, entails evaluating the recurrence

\begin{equation*}
\left\{\begin{array}{ll}
	\vv{h}_1 = W_1^T\vv{\phi} + \vv{b}_1 & \textrm{from input to first hidden layer}, \\
	\vv{h}_{i+1} = \textrm{ReLU}(W_{i+1}^T\vv{h}_i + \vv{b}_{i+1}), \; 1 \leqslant i \leqslant 3 & \textrm{from the }i\textrm{th to the }(i+1)\textrm{th hidden layers}, \\
	h\kappa = W_5^T\vv{h}_4 + \vv{b}_5 & \textrm{from last hidden to output layer,}
\end{array}\right.
\end{equation*}
where $W_1 \in \mathbb{R}^{9\times N_h^1}$, $W_{i+1} \in \mathbb{R}^{N_h^i\times N_h^{i+1}}$, and $W_5 \in \mathbb{R}^{N_h^4\times 1}$ are $F_h(\cdot)$'s weight matrices, and $\vv{b}_m$ are its corresponding bias vectors, for $1 \leqslant m \leqslant 5$.

Our choice for rectified linear units is based on their benefits for feedforward networks compared to other traditional sigmoidal neurons.  Although the function $\textrm{ReLU}(z) = \max(0, z)$ is non-differentiable at $z=0$, it works well in practice with gradient descent because it is fast to compute and does not saturate for $z>0$ \cite{Hands-onMLwithScikit-LearnKerasAndTF19}.  In particular, it is assumed that $\textrm{ReLU}'(z)=0$ for $z \leqslant 0$, while $\textrm{ReLU}'(z)=1$ for $z > 0$.  Also, the ReLU's non-saturating feature favors gradient propagation on the active path of neurons, thus circumventing almost entirely the vanishing-gradient problem with no need for unsupervised pre-training \cite{DeepSparseRectifierNN;Glorot-et-al;2011}.  Furthermore, ReLUs are very close to being linear and enjoy many of the properties that make linear models easy to optimize with gradient descent \cite{DeepLearning;Goodfellow-et-al;2016}.  For a thorough analysis of rectified linear neurons in backpropagation, we refer the interested reader to the groundbreaking study by Glorot \etal \cite{DeepSparseRectifierNN;Glorot-et-al;2011}; similarly, Chapter 11 in \cite{Hands-onMLwithScikit-LearnKerasAndTF19} illustrates the use of ReLUs, their properties, and their variants in Keras-based applications.

Here, we have used TensorFlow's backpropagation and the Adam optimizer \cite{Adam;2015} to adjust the connection weights and biases and minimize the mean squared error loss function (MSE) computed off the outputs in $\mathcal{D}$'s training subset.  At the beginning, the model weights are set with a Glorot uniform initializer \cite{Glorot;Bengio;2010}, and the biases start with zero values \cite{Keras15}.  Then, every parameter update takes place after observing a batch of 64 samples.  Moreover, we monitor the validation subset mean absolute error metric (MAE) for two primary reasons.  First, it allows us to stabilize the optimization process by halving the learning rate (starting at $0.00015$) whenever the MAE does not improve for fifteen epochs.  Second, it helps us prevent overfitting by stopping backpropagation if the MAE stagnates for sixty epochs.  These aspects, among others, are integrated into the fitting program using Keras' built-in callback functions.  As a final comment, we remark that the hyperparameters just provided have been found to generalize well, irrespective of the domain resolutions studied in Section \ref{sec:ExperimentsAndResults}.  As for choosing the size of the hidden layers, it largely depends on $h$ and the quality of the approximated distances to $\Gamma$ in $\vv{\phi}$.  For compactness, we have omitted the mathematical details underlying backpropagation and stochastic optimization using Adam; however, we refer the interested reader to \cite{A18} and \cite{Tensorflow15, Keras15, Adam;2015} for a thorough description of these important machine learning aspects.


\colorsection{Experiments and results}
\label{sec:ExperimentsAndResults}

Now, we introduce a few performance tests of the hybrid inference system over stationary, smooth interfaces.  We begin by providing a detailed comparison between our framework and the compound numerical method (i.e., $G_h(\cdot)$) in a coarse adaptive grid.  This assessment precedes a couple of convergence tests where we consider increasingly higher local resolutions.

The first two experiments evaluate the accuracy of dimensionless curvatures estimated at the zero level set of the bivariate function

\begin{equation}
	\phi_{rose}(x, y) = \sqrt{x^2 + y^2} - a\cos(p\theta) - b,
	\label{eq:polarRose}
\end{equation}
where $\theta \in [0, 2\pi)$ is the angle of the vector toward $(x, y)$ with respect to the horizontal, and $a, b, p$ are shape parameters.  The interface $r(\theta) = a\cos(p\theta) + b$ is commonly referred to in the literature as the ``polar rose.''  One can determine the curvature along $\Gamma$ for this simple form with the expression

\begin{equation}
\kappa_{rose}(\theta) = \frac{r^2(\theta) + 2\left(r'(\theta)\right)^2 - r(\theta)r''(\theta)}{\left(r^2(\theta) + \left(r'(\theta)\right)^2\right)^{3/2}}.
\label{eq:kappaPolarRose}
\end{equation}

In these initial experiments, we set $p = 5$ and vary $a$ and $b$ to change the steepness of the five petal junctions.  Finally, we conclude this section with a circle-based convergence test, comparing our neural system with the hybrid particle-VOF method described in \cite{Karnakov;etal;HybridParticleMthdVOFCuravture;2020}.  For all cases, we reiterate that we have computed the training and testing samples from reinitialized level-set functions using ten iterations to solve \eqref{eq:reinitialization}.


\colorsubsection{Evaluation in a low-resolution grid}
\label{subsec:EvaluationAtLowResolution}

In our prior research \cite{LALariosFGibou;LSCurvatureML;2021}, we hinted at incorporating machine learning for helping the level-set method to handle under-resolved regions and steep curvatures in low resolutions.  Here, we realize that idea and apply our novel hybrid inference system to estimating dimensionless curvature in a relatively coarse adaptive grid.

Let $\Omega$ be a non-periodic computational domain discretized into an adaptive \verb|p4est| macromesh \cite{Strain1999, Burstedde;Wilcox;Ghattas:11:p4est:-Scalable-Algo}.  Further, let $\ell_{max} \in \mathbb{N}$ be the maximum level of refinement of each of the $1\times 1$ macrocells or quadtrees \cite{BKOS00} that compose $\Omega$.  The recursive quadtree cells' subdivision down to the finest local resolution abides by the splitting criterion proposed by \cite{Min;Gibou:07:A-second-order-accur},

\begin{equation}
\min_{v\in\mathcal{V}(\mathcal{c})}|\phi(v)| \leqslant \textrm{Lip}\left(\phi(v)\right) \times \textrm{diag}_\ell(\mathcal{c}),
\label{eq:quadtreeRefinementCriterion}
\end{equation}
where $\mathcal{V}$ is the set of vertices in cell $\mathcal{c}$, $\textrm{diag}_\ell(\mathcal{c})$ is its diagonal length, and $\textrm{Lip}(\phi(v))$ is the Lipschitz constant for the level-set function (set to 1.2 for all $\mathcal{c}$).

\begin{figure}[!t]
	\centering
	\begin{subfigure}[b]{0.35\textwidth}
		\includegraphics[width=\textwidth]{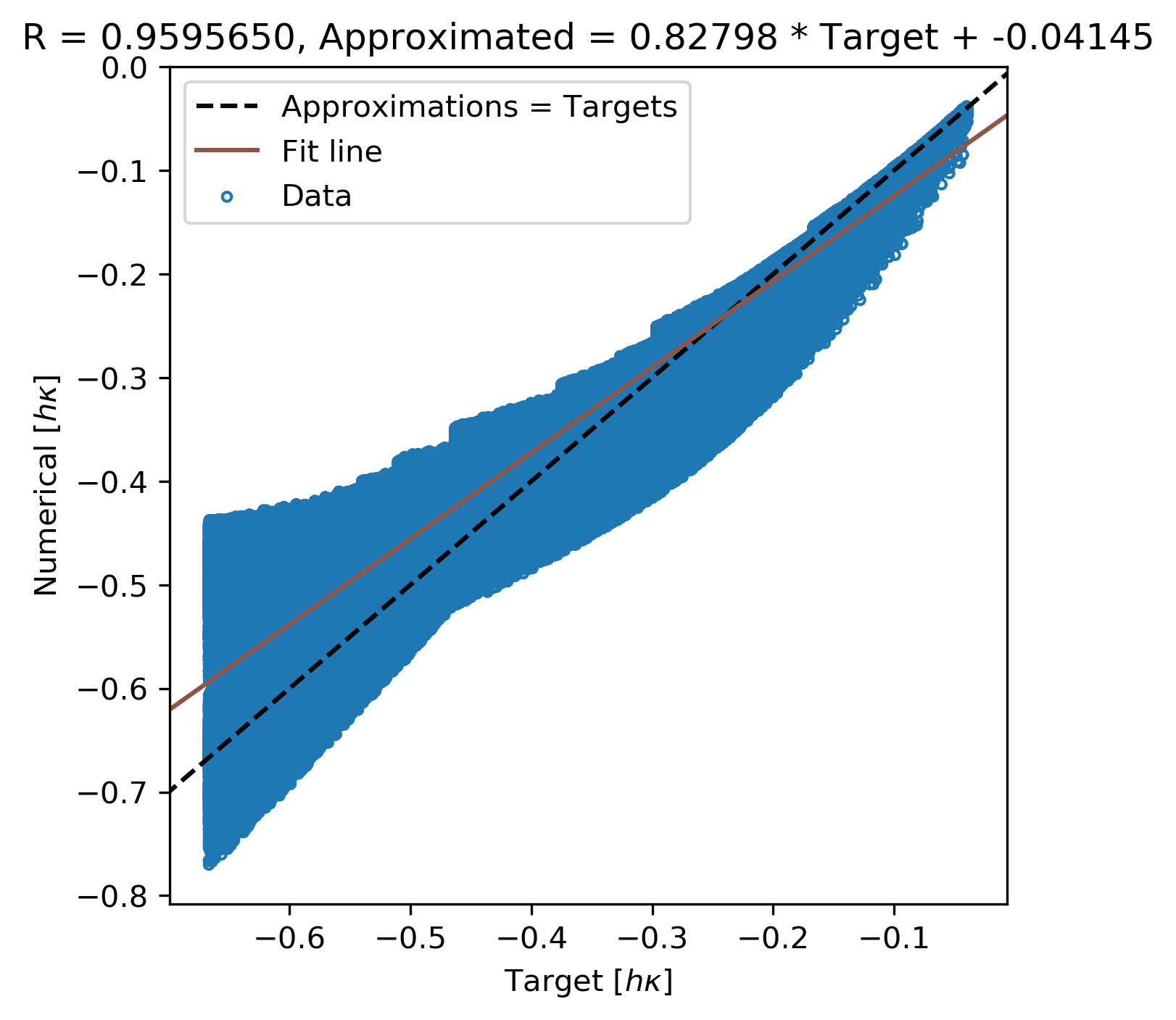}
        \caption{\footnotesize Compound numerical method}
        \label{fig:models.7.hybrid.numerics}
    \end{subfigure}
    ~
	\begin{subfigure}[b]{0.35\textwidth}
		\includegraphics[width=\textwidth]{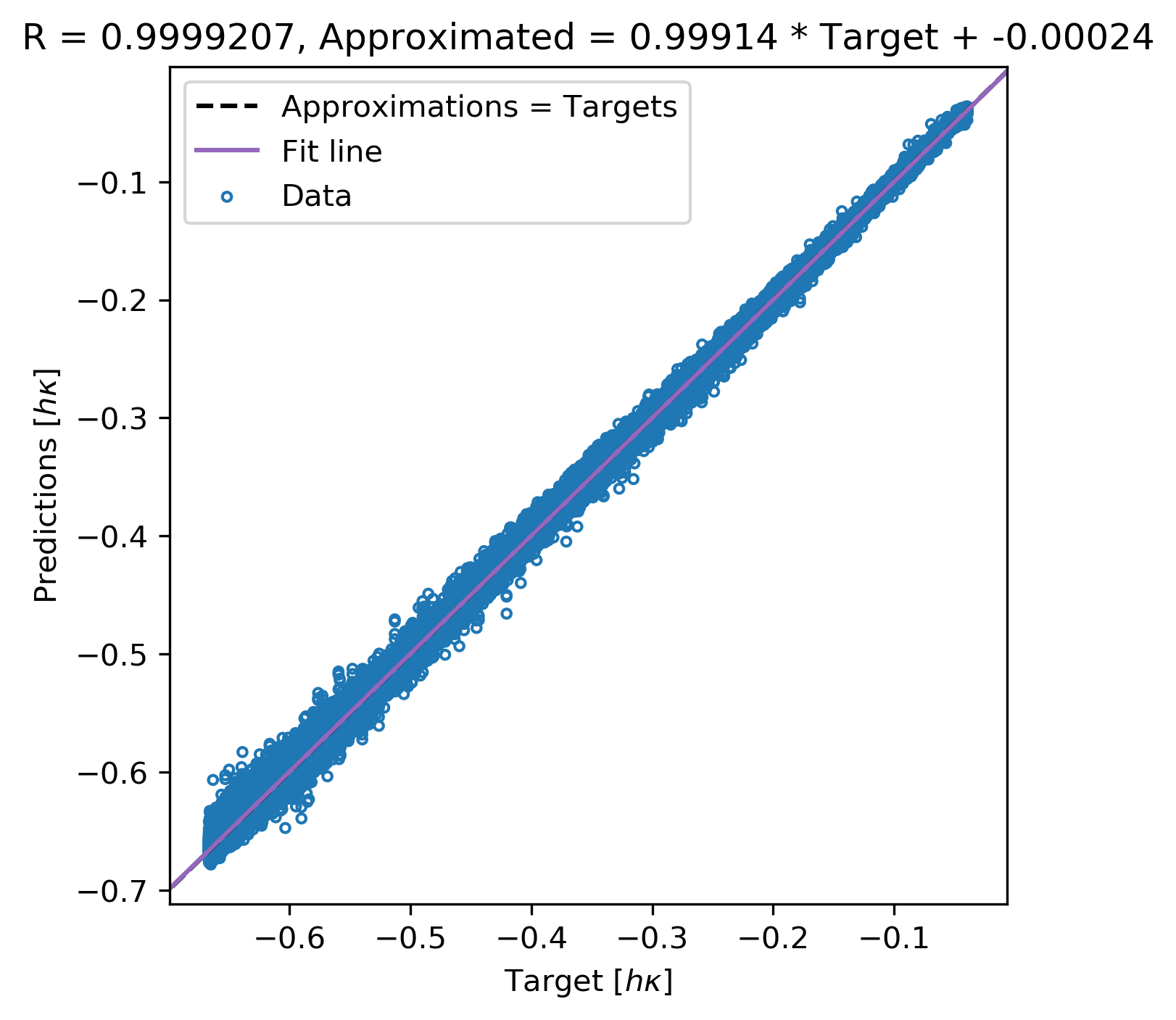}
        \caption{\footnotesize Neural network}
        \label{fig:models.7.hybrid.reinit}
    \end{subfigure}
    
	\caption{Training correlation plots between expected and approximated $h\kappa$ over $\mathcal{D}_{rls}$ in an adaptive grid with $\ell_{max} = 7$.  (Color online.)}
	\label{fig:models.7.hybrid}
\end{figure}

By defining $\ell_{max} = 7$, we get a minimum, uniform local mesh size of $h = 1/128$.  We have trained a suitable neural network, $F_{hyb}^{(7)}(\cdot)$, for this spacing with 2'330,785 samples extracted from sinusoidal- and circular-interface level-set functions in the negative curvature spectrum, $-85\frac{1}{3} \leqslant \kappa \leqslant -\frac{1}{2}$.  The optimal multilayer perceptron for this case has four hidden layers with 180 ReLU units each (refer to Figure \ref{fig:mlp}).  Such an architecture adds up to 99,721 parameters among weight matrices and biases.  Figure \ref{fig:models.7.hybrid.reinit} compares the fitting quality of $F_{hyb}^{(7)}(\cdot)$ and the compound numerical method, $G_h(\cdot)$, over $\mathcal{D}_{rls}$.  This data set contains pairs $(\vv{\phi}, h\kappa)$, where $\vv{\phi}$ is a stencil of reinitialized level-set values.  Also, note that we used non-overlapping portions of $\mathcal{D}_{rls}$ in the training, testing, and validation subsets of $\mathcal{D}$.  Figure \ref{fig:models.7.hybrid.reinit} shows the training outcome by considering those samples that simulate a practical scenario where only reinitialized level-set information is available.  A quick inspection of the plots in Figure \ref{fig:models.7.hybrid} reveals the superior accuracy of our regression model and confirms that machine learning works better when $|\kappa|$ is large.  Table \ref{tbl:models.7.hybrid.stats} supplements these results with some error statistics over $\mathcal{D}_{rls}$ regarding raw curvature values.  The difference between both methods is surely drastic.

\begin{table}[!b]
	\centering
	\small
	\bgroup
	\def\arraystretch{1.1}%
	\begin{tabular}{|l|r|r|r|r|}
		\hline
		~ & Mean absolute error & Maximum absolute error & Mean squared error & Training epochs \\
		\hline \hline
		Neural network   & $\eten{1.815692}{-1}$ &           $7.242649$ & $\eten{8.291462}{-2}$ & 434 \\ \hline
		Numerical method &            $4.511332$ & $\eten{2.926552}{1}$ &  $\eten{5.141757}{1}$ & - \\
		\hline
	\end{tabular}
	\egroup
	\caption{Training statistics with respect to $\kappa$ over $\mathcal{D}_{rls}$ in an adaptive grid with $\ell_{max} = 7$.}
	\label{tbl:models.7.hybrid.stats}
\end{table}

\begin{figure}[!b]
	\centering
	\begin{subfigure}[b]{0.35\textwidth}
		\includegraphics[width=\textwidth]{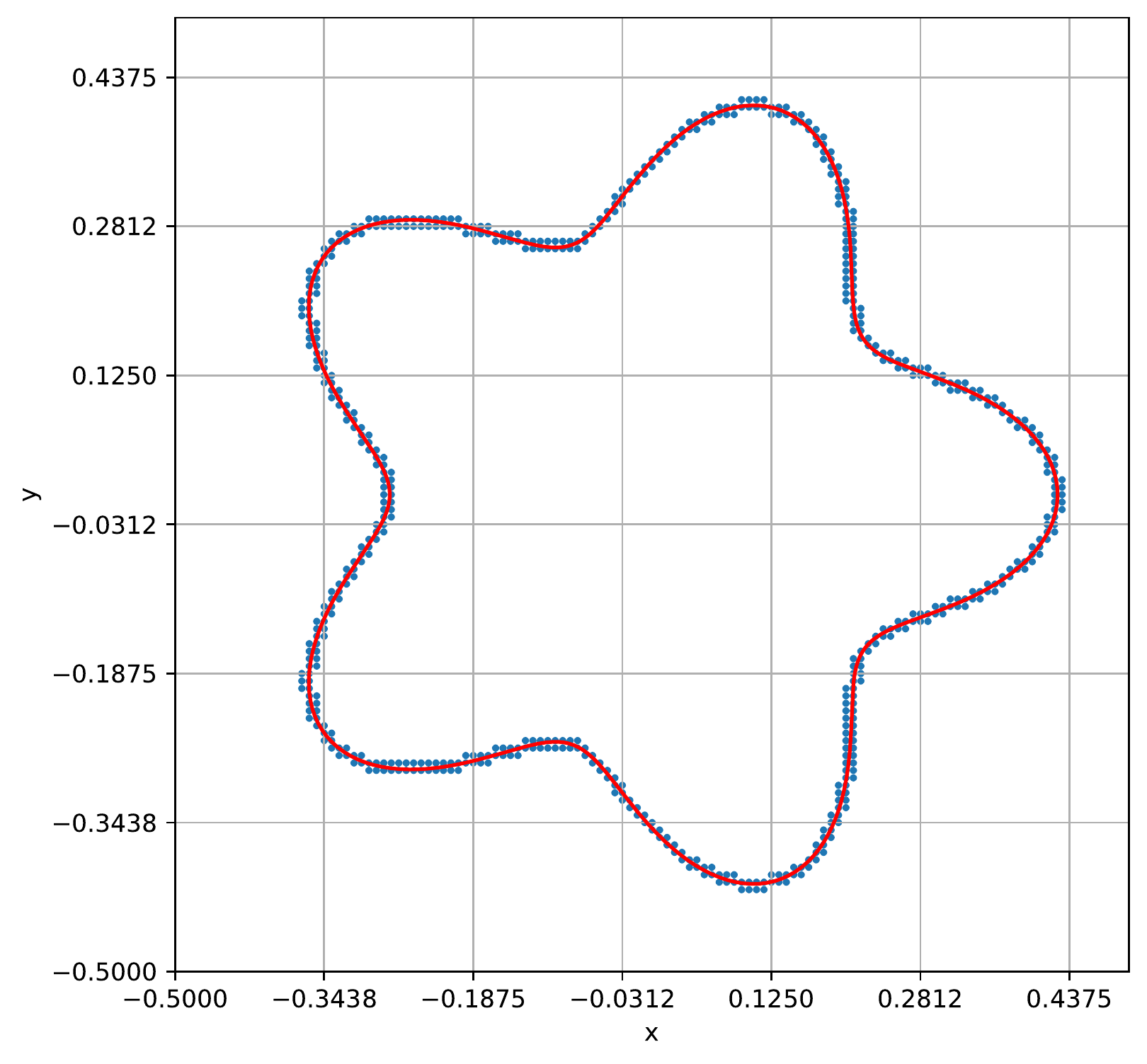}
		\caption{\footnotesize $\Gamma_1$: $r(\theta) = 0.075\sin(5\theta) + 0.35$}
		\label{fig:results.7.hybrid.rose.a075.b350}
	\end{subfigure}
	~
	\begin{subfigure}[b]{0.35\textwidth}
		\includegraphics[width=\textwidth]{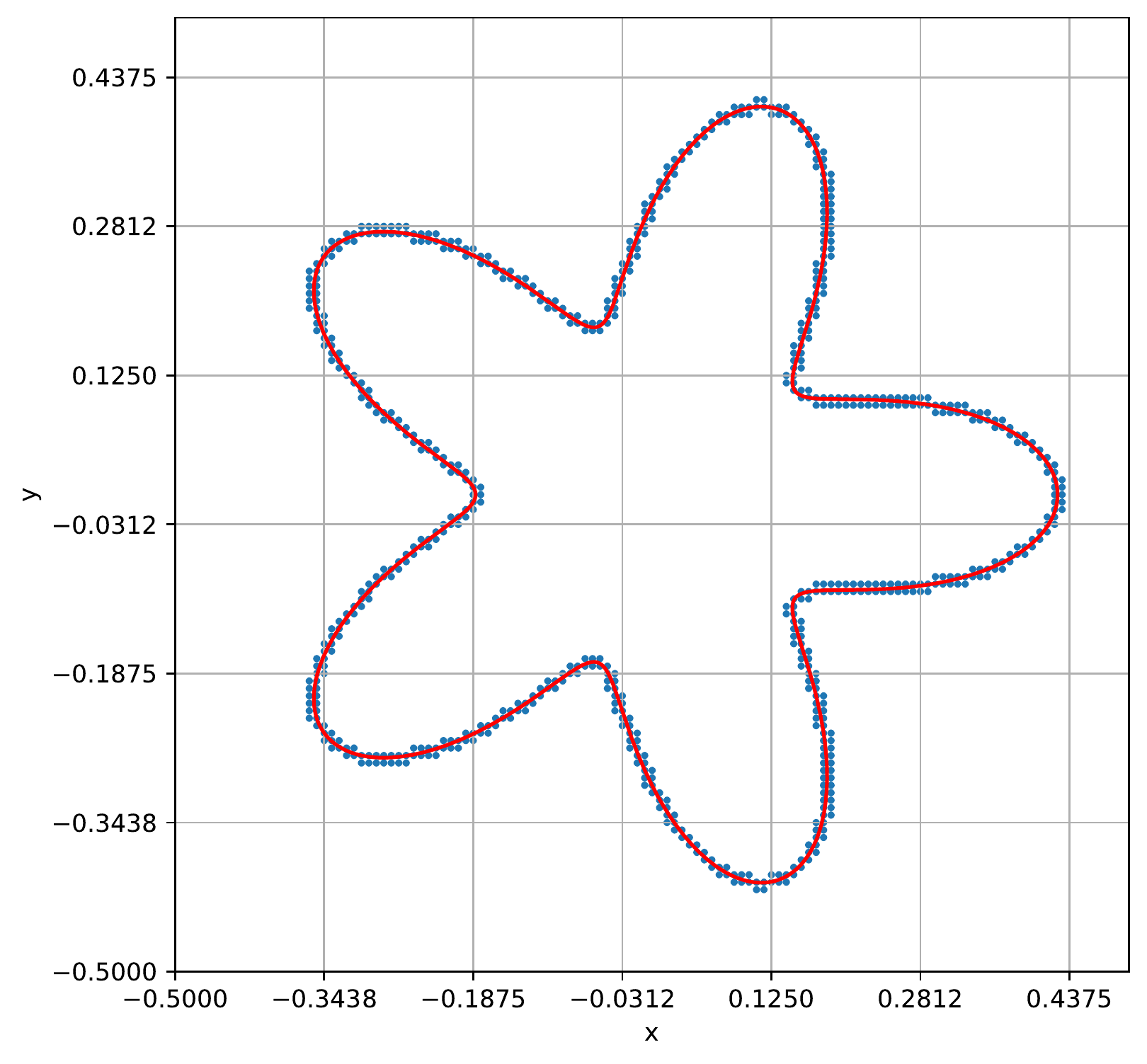}
		\caption{\footnotesize $\Gamma_2$: $r(\theta) = 0.12\sin(5\theta) + 0.305$}
		\label{fig:results.7.hybrid.rose.a120.b305}
	\end{subfigure}
	\caption{Two shape configurations of the polar rose interface.  The continuous red line denotes the interface, and the blue dots indicate samples collected next to $\Gamma$.  (Color online.)}
	\label{fig:results.7.hybrid.rose}
\end{figure}

We first test $F_{hyb}^{(7)}$ and the hybrid inference system of Algorithm \ref{alg:hybridApproach} on $\phi_{rose}(x, y)$.  Let $a = 0.075$ and $b = 0.35$.  These shape parameters define a five-petaled $C^1$ interface, $\Gamma_1$, as seen in Figure \ref{fig:results.7.hybrid.rose.a075.b350}.  Then, we discretize the square computational domain $\Omega \equiv [-0.5, +0.5]^2$ with a single quadtree with $\ell_{max} = 7$, which produces a minimum spacing of $h = 1/128$.  These settings allow us to gather 632 data pairs for which we can calculate their curvature exactly with \eqref{eq:kappaPolarRose} at their closest points on $\Gamma_1$.  Similar to locating nearest points to vertices along sine waves (Section \ref{subsec:SinusoidalInterfaceDataSetGeneration}), we use bisection and Newton--Raphson's root finding to determine the optimal angle, $\theta^*$, that yields $\vv{x}_{i, j}^\perp$ in Figure \ref{fig:stencil}.  In addition, we keep the network threshold $\kappa_{flat} = 5$, as stated at the end of Section \ref{subsec:HybridInferenceSystem}.

\begin{figure}[t]
	\centering
	\begin{subfigure}[b]{0.32\textwidth}
		\includegraphics[width=\textwidth]{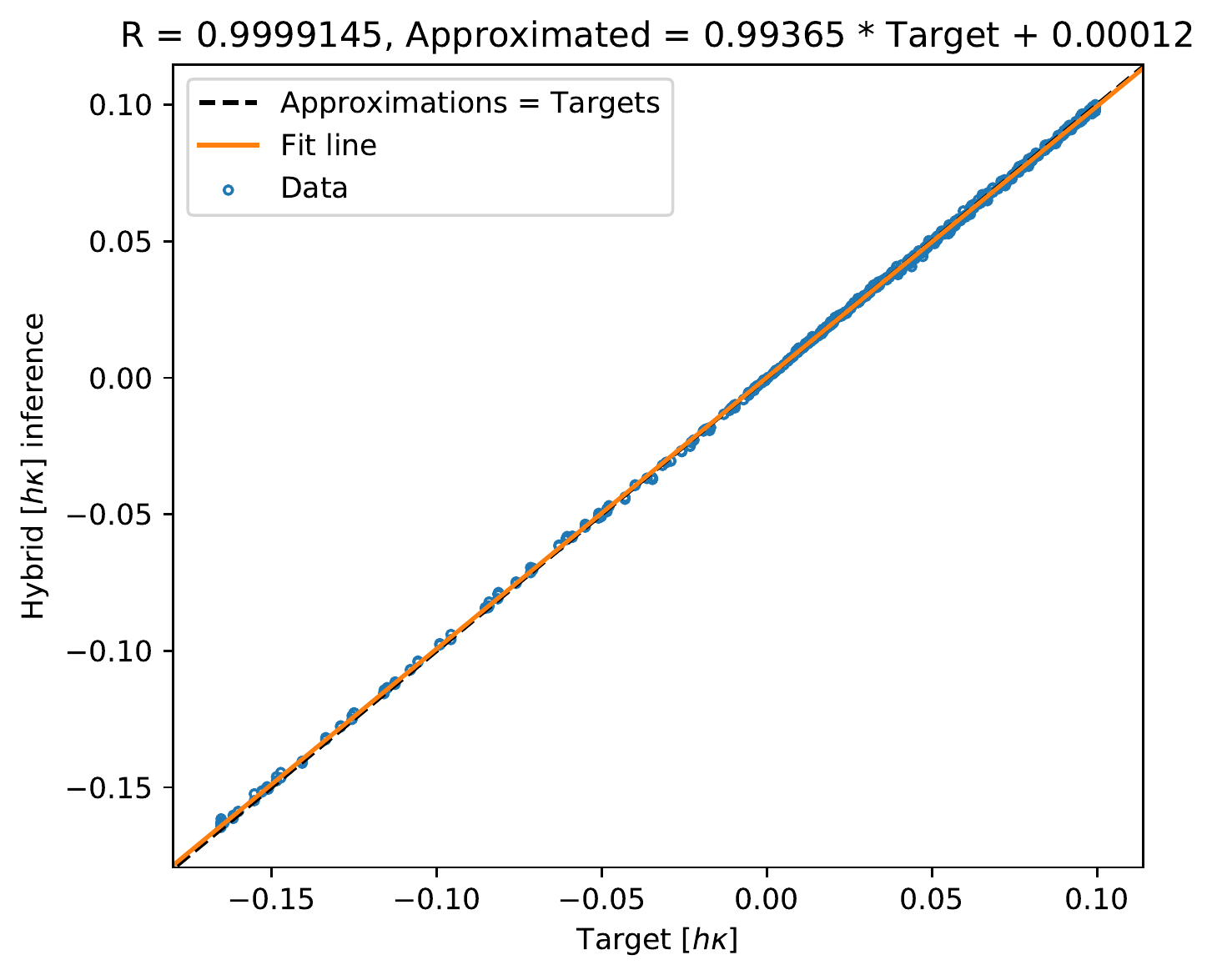}
        \caption{\footnotesize Hybrid inference}
        \label{fig:results.7.hybrid.rose.a075.b350.nnet}
    \end{subfigure}
	\begin{subfigure}[b]{0.32\textwidth}
		\includegraphics[width=\textwidth]{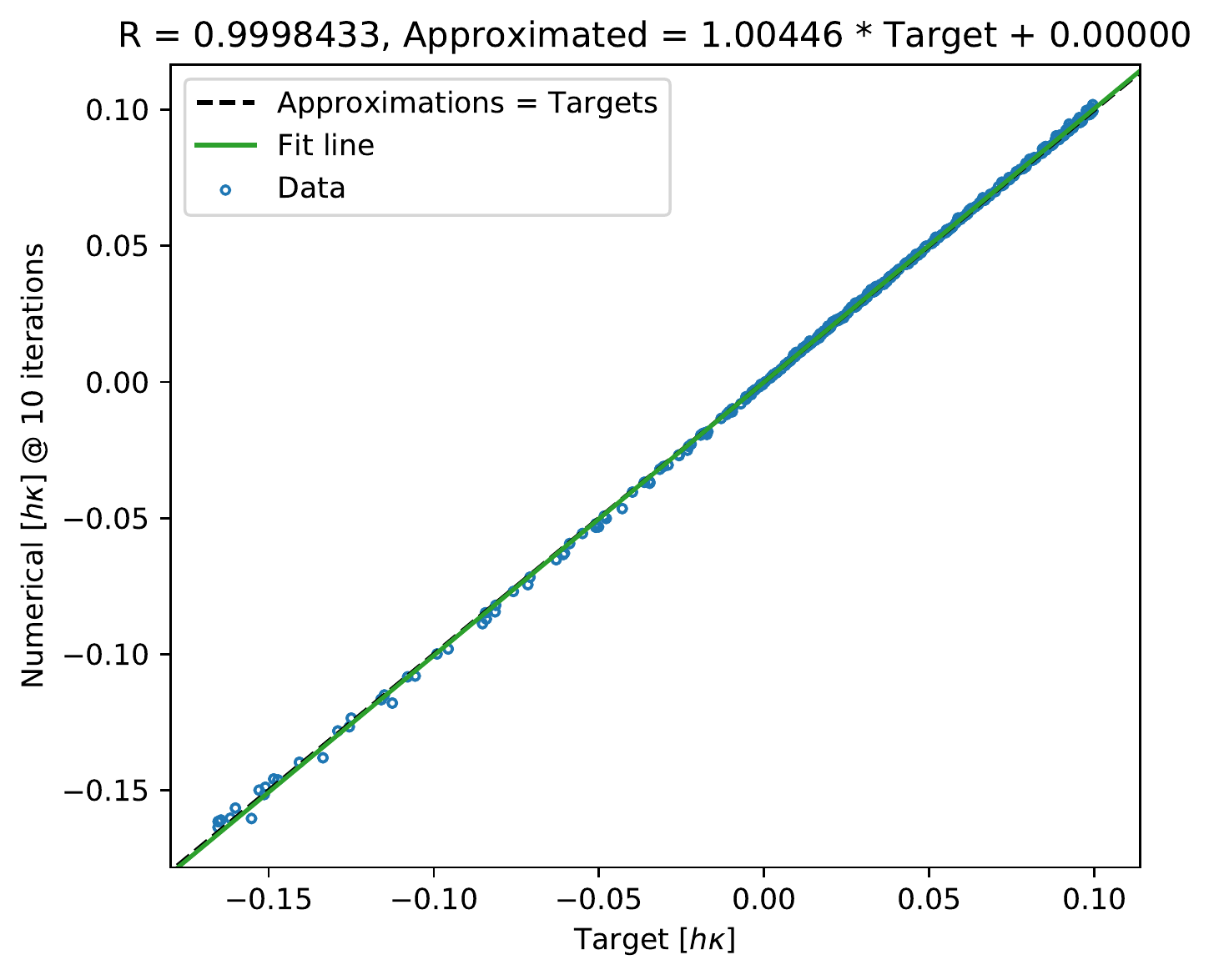}
		\caption{\footnotesize Numerical, 10 iterations}
		\label{fig:results.7.hybrid.rose.a075.b350.numerical.10}
	\end{subfigure}
	\begin{subfigure}[b]{0.32\textwidth}
		\includegraphics[width=\textwidth]{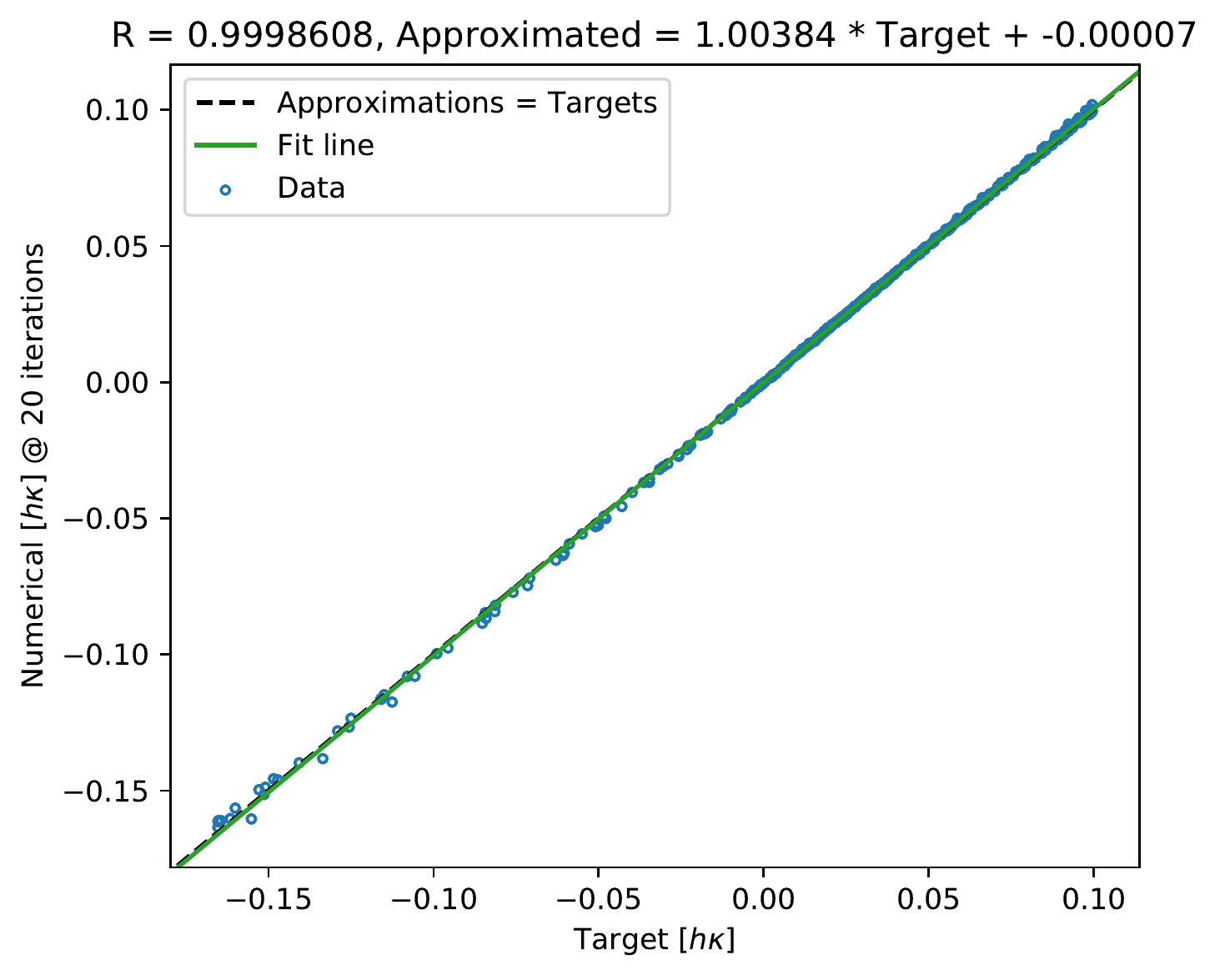}
		\caption{\footnotesize Numerical, 20 iterations}
		\label{fig:results.7.hybrid.rose.a075.b350.numerical.20}
	\end{subfigure}
   
	\caption{Correlation between expected and approximated $h\kappa$ using the hybrid inference system and the compound numerical method on $\Gamma_1$ in an adaptive grid with $\ell_{max} = 7$.  (Color online.)}
	\label{fig:results.7.hybrid.rose.a075.b350.correlation}
\end{figure}

Figure \ref{fig:results.7.hybrid.rose.a075.b350.correlation} illustrates the quality of the dimensionless curvature estimations computed with our hybrid approach and the compound numerical method.  For reference, we also include the numerical approximations when we reinitialize $\phi_{rose}$ with 20 iterations.  It is then easy to see that our strategy delivers comparable or better results than the conventional numerical schemes when the interface features slowly varying curvatures.  Table \ref{tbl:results.7.hybrid.rose.a075.b350.summary} has more to say about this.  It provides $\kappa$ error statistics and shows that our framework is slightly more accurate in the $L^\infty$ norm.  The latter holds even if one uses $G_h(\cdot)$ alone with twice the number of redistancing operations necessary for training $F_{hyb}^{(7)}$.  Lastly, Table \ref{tbl:results.7.hybrid.rose.a075.b350.summary} tells us that, on average, the numerical method gets better at estimating $\kappa$ as the number of iterations increases.  This behavior is expected, but the gained advantage is not significant if we contrast it with the MAE incurred by the hybrid algorithm.

\begin{table}[!b]
	\centering
	\small
	\bgroup
	\def\arraystretch{1.1}%
	\begin{tabular}{|l|r|r|r|}
		\hline
		Method & Mean absolute error & Maximum absolute error & Mean squared error \\
		\hline \hline
		Hybrid approach             & $\eten{8.559648}{-2}$ & $\eten{4.765949}{-1}$ & $\eten{1.298467}{-2}$ \\ 
		\hline
 		Numerical method, 10 iters. & $\eten{9.129319}{-2}$ & $\eten{6.847077}{-1}$ & $\eten{2.099346}{-2}$ \\
		\hline
 		Numerical method, 20 iters. & $\eten{7.665316}{-2}$ & $\eten{6.701565}{-1}$ & $\eten{1.833753}{-2}$ \\
		\hline
	\end{tabular}
	\egroup
	\caption{Error analysis with respect to $\kappa$ for $\Gamma_1$ in an adaptive grid with $\ell_{max} = 7$.}
	\label{tbl:results.7.hybrid.rose.a075.b350.summary}
\end{table}

\begin{figure}[t]
	\centering
	\begin{subfigure}[b]{0.32\textwidth}
		\includegraphics[width=\textwidth]{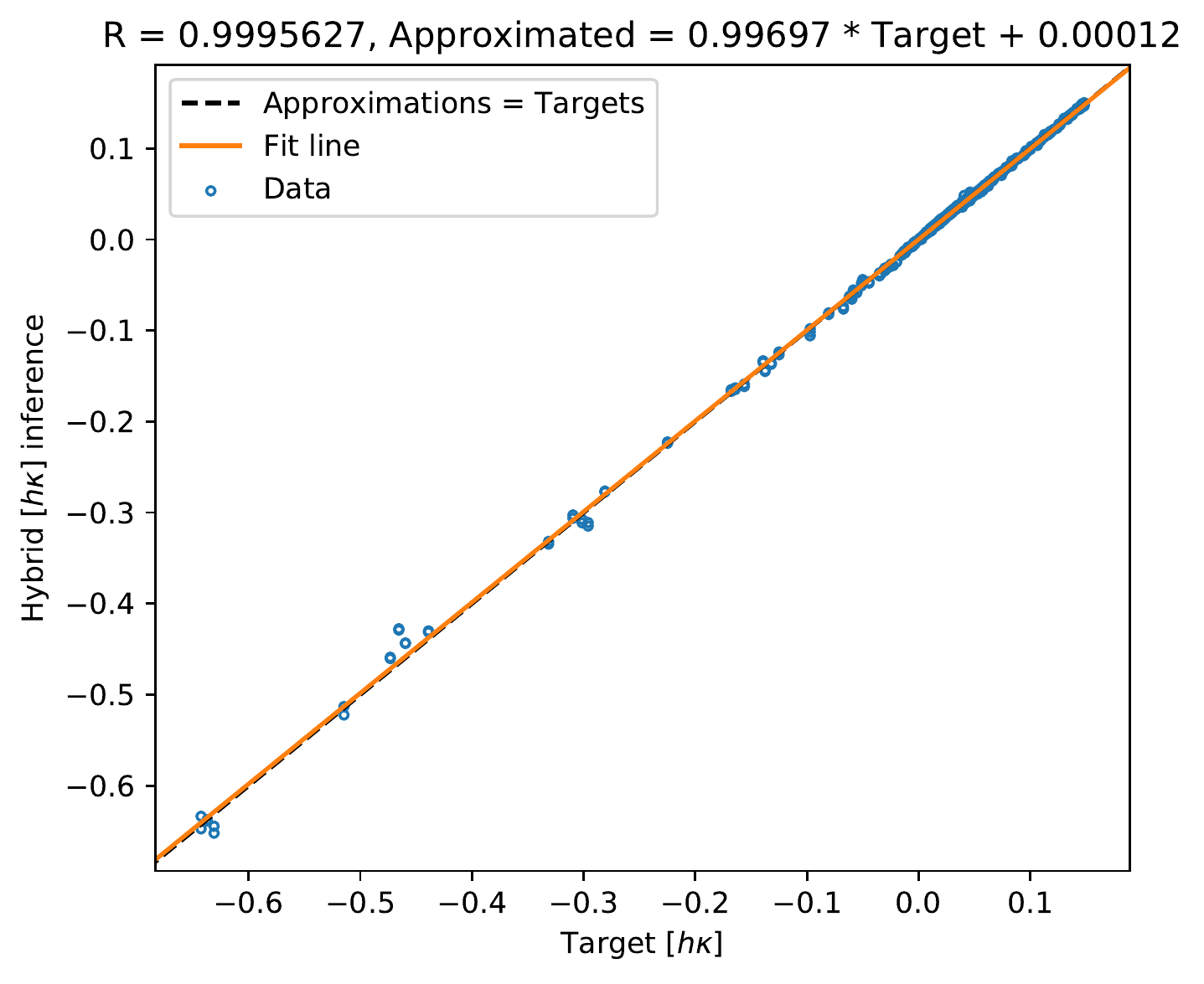}
        \caption{\footnotesize Hybrid inference}
        \label{fig:results.7.hybrid.rose.a120.b305.nnet}
    \end{subfigure}
	\begin{subfigure}[b]{0.32\textwidth}
		\includegraphics[width=\textwidth]{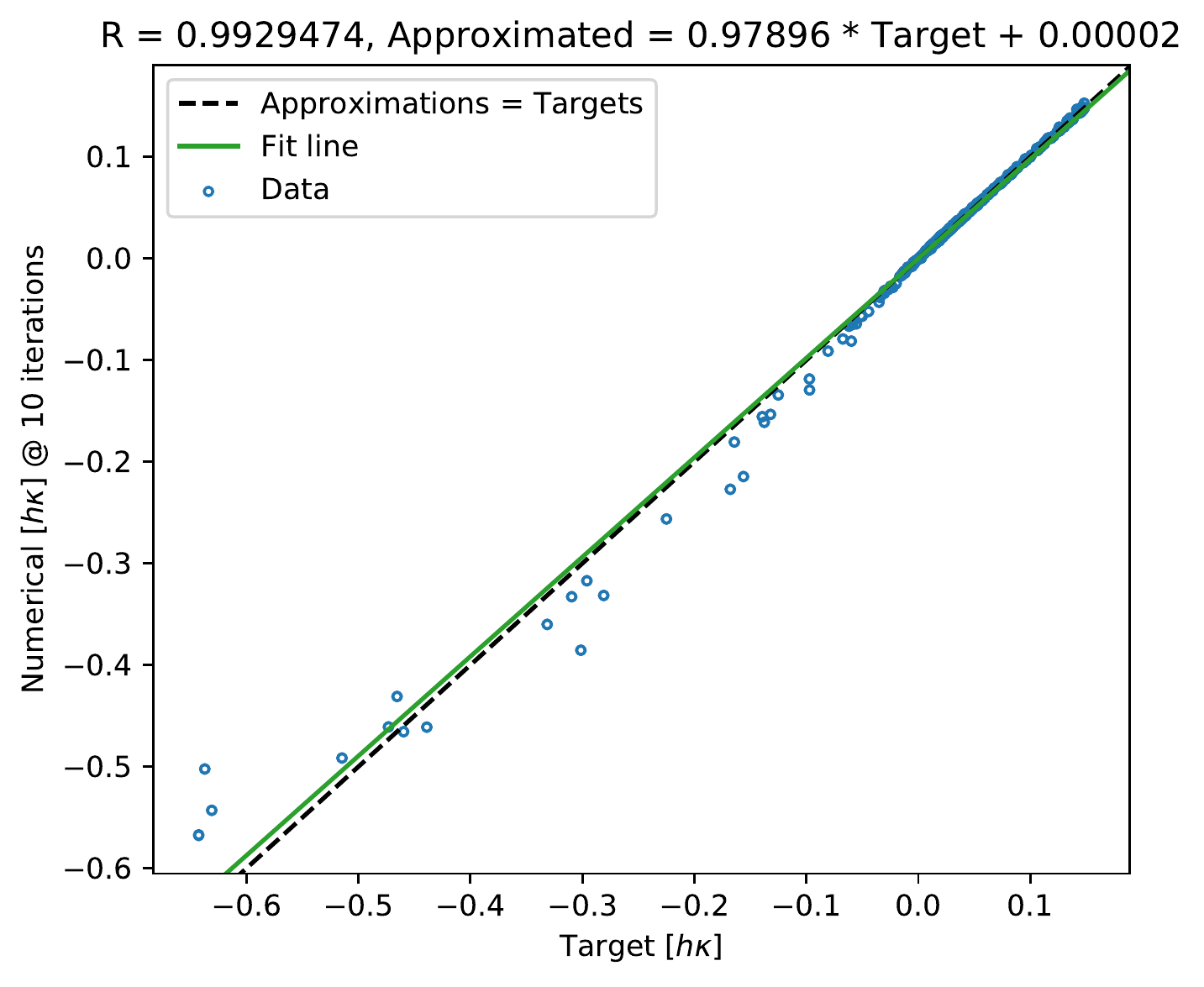}
		\caption{\footnotesize Numerical, 10 iterations}
		\label{fig:results.7.hybrid.rose.a120.b305.numerical.10}
	\end{subfigure}
	\begin{subfigure}[b]{0.32\textwidth}
		\includegraphics[width=\textwidth]{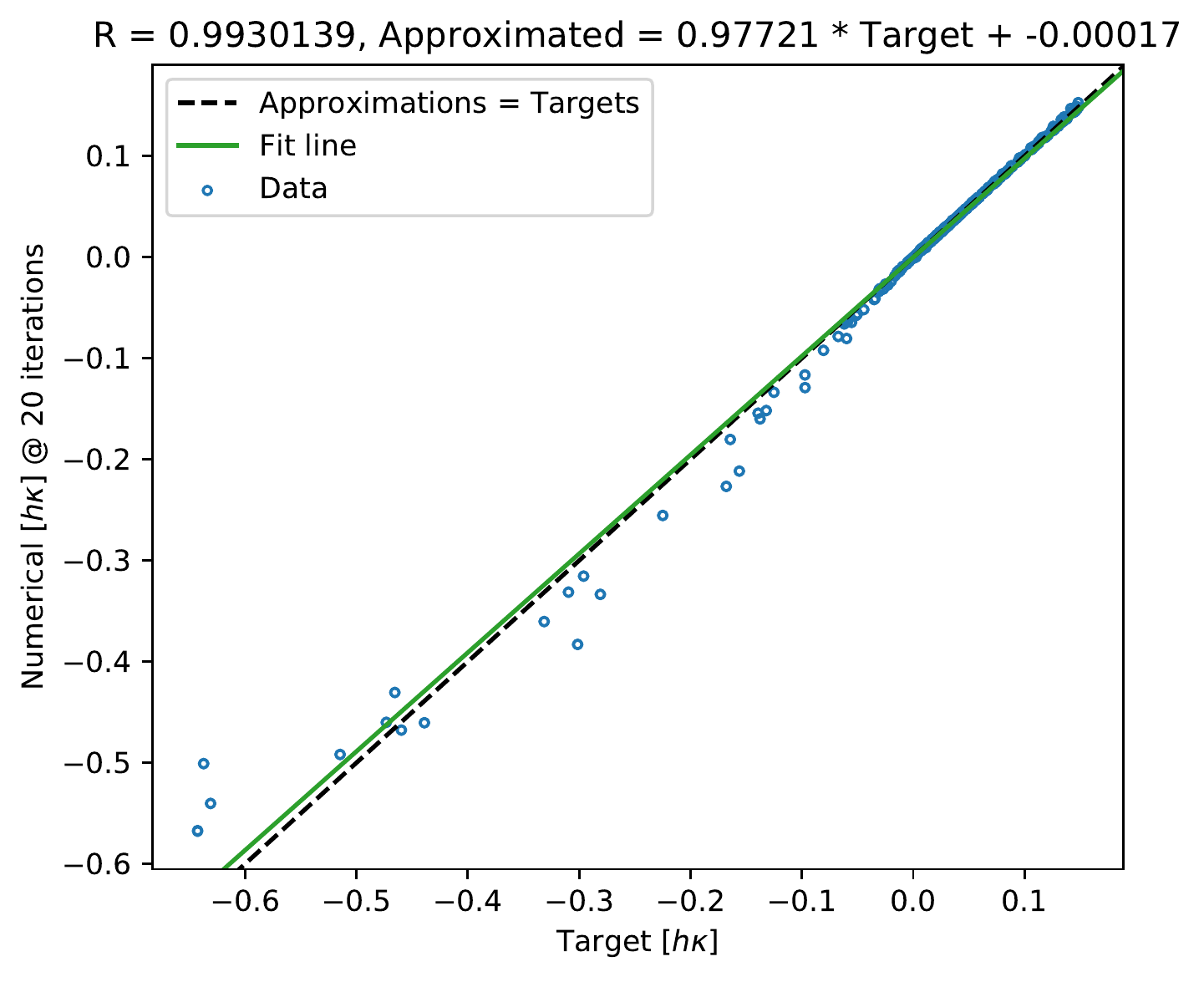}
		\caption{\footnotesize Numerical, 20 iterations}
		\label{fig:results.7.hybrid.rose.a120.b305.numerical.20}
	\end{subfigure}
   
	\caption{Correlation between expected and approximated $h\kappa$ using the hybrid inference system and the compound numerical method on $\Gamma_2$ in an adaptive grid with $\ell_{max} = 7$.  (Color online.)}
	\label{fig:results.7.hybrid.rose.a120.b305.correlation}
\end{figure}

In a second test, we consider the function in \eqref{eq:polarRose} with $a = 0.12$ and $b = 0.305$.  This configuration is characterized by a zero level set, $\Gamma_2$, that displays steeper concave regions than $\Gamma_1$, as seen in Figure \ref{fig:results.7.hybrid.rose.a120.b305}.  The experiment setup is similar to the previous case, except that the number of samples collected along the interface is 740.  Figure \ref{fig:results.7.hybrid.rose.a120.b305.correlation} shows the correlation plots obtained with our hybrid approach and the compound numerical method.  For reference, we include again the numerical estimations attained with 20 iterations for redistancing $\phi_{rose}(x,y)$.

Comparing Figures \ref{fig:results.7.hybrid.rose.a075.b350.correlation} and \ref{fig:results.7.hybrid.rose.a120.b305.correlation} reveals that $\Gamma_2$ is more challenging to deal with, given the relatively coarse mesh discretization.  Both methods struggle, but the hybrid inference system is by far superior at handling steep curvatures regardless of the number of iterations used to reinitialize the level-set function.  Table \ref{tbl:results.7.hybrid.rose.a120.b305.summary} summarizes these results more succinctly.  It shows that the hybrid strategy is up to 3.5 times more accurate in the $L^\infty$ norm and incurs as much as 44\% of $G_h(\cdot)$'s MAE.  Unlike \cite{LALariosFGibou;LSCurvatureML;2021}, the gains in robustness and correctness are significant when one integrates a greater diversity of training samples to optimize a curvature regression model.

\begin{table}[!b]
	\centering
	\small
	\bgroup
	\def\arraystretch{1.1}%
	\begin{tabular}{|l|r|r|r|}
		\hline
		Method & Mean absolute error & Maximum absolute error & Mean squared error \\
		\hline \hline
		Hybrid approach             & $\eten{1.778957}{-1}$ &           $4.841265$ & $\eten{1.601629}{-1}$ \\ 
		\hline
 		Numerical method, 10 iters. & $\eten{4.441347}{-1}$ & $\eten{1.726560}{1}$ &            $2.570065$ \\
		\hline
 		Numerical method, 20 iters. & $\eten{4.046386}{-1}$ & $\eten{1.745362}{1}$ &            $2.554465$ \\
		\hline
	\end{tabular}
	\egroup
	\caption{Error analysis with respect to $\kappa$ for $\Gamma_2$ in an adaptive grid with $\ell_{max} = 7$.}
	\label{tbl:results.7.hybrid.rose.a120.b305.summary}
\end{table}


\colorsubsection{Flower-shaped-interface-based convergence study}
\label{subsec:FlowerShapeConvergenceStudy}

We now study the behavior of the hybrid framework as $h \rightarrow 0$ on the irregular interface $\Gamma_2$.  We have chosen this interface because of its challenging steep concave regions.  To measure the hybrid strategy performance objectively, we compare it with each of its components alone.  More clearly, we consider a network-only inference system (i.e., $P_h(\cdot) + F_h(\cdot)$, as in \cite{CurvatureML19, LALariosFGibou;LSCurvatureML;2021, VOFCurvature3DML19}), the conventional compound numerical method (i.e., $G_h(\cdot)$), and the hybrid solver in Algorithm \ref{alg:hybridApproach}.

The convergence analysis realizes the idea of building independent neural networks for each grid resolution.  This study agrees with the thesis that a dictionary of neural models is simpler and more accurate than a universal regressor.  Therefore, we have trained suitable models, $F_{net}^{(\nu)}$ and $F_{hyb}^{(\nu)}$, for the network-only and hybrid strategies from sinusoidal- and circular-interface samples in four resolutions, $h_\nu = 2^{-\nu}$, where $\nu = 7, 8, 9, 10$.  To assemble the corresponding learning data sets $\mathcal{D}^{(\nu)}$, we consistently use $\kappa_{min} = 0.5$, $\kappa_{max} = 85\frac{1}{3}$, and $\kappa_{flat} = 5$, as described in Sections \ref{subsec:CircularInterfaceDataSetGeneration} and \ref{subsec:SinusoidalInterfaceDataSetGeneration}.  Regarding Algorithm \ref{alg:hybridApproach}, however, we opt for redefining the inference threshold $\kappa_{flat}^{(\nu)} = 2^{(\nu - 7)}\kappa_{flat}$,  so that the solver prefers $G_h(\cdot)$ over $F_h(\cdot)$ in wider curvature intervals around 0, as $\nu \rightarrow 10$.  The goal is to leverage the traditional numerical framework, knowing that its accuracy increases as the mesh becomes finer.

\begin{figure}[t]
	\centering
	\begin{subfigure}[b]{0.32\textwidth}
		\includegraphics[width=\textwidth]{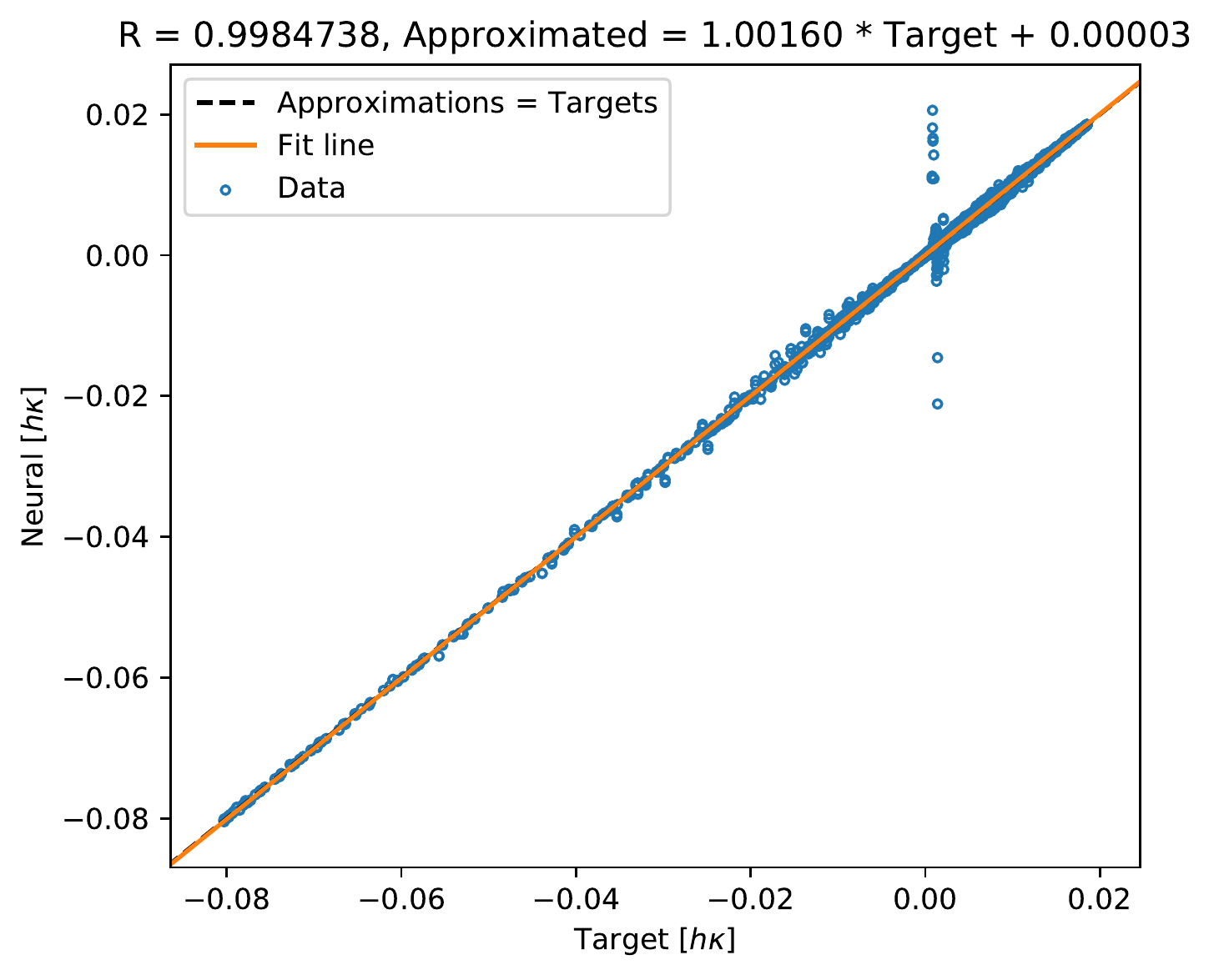}
        \caption{\footnotesize Using PCA and whitening}
        \label{fig:pathological.10.full.rose.a120.b305.nnet}
    \end{subfigure}
    ~
	\begin{subfigure}[b]{0.32\textwidth}
		\includegraphics[width=\textwidth]{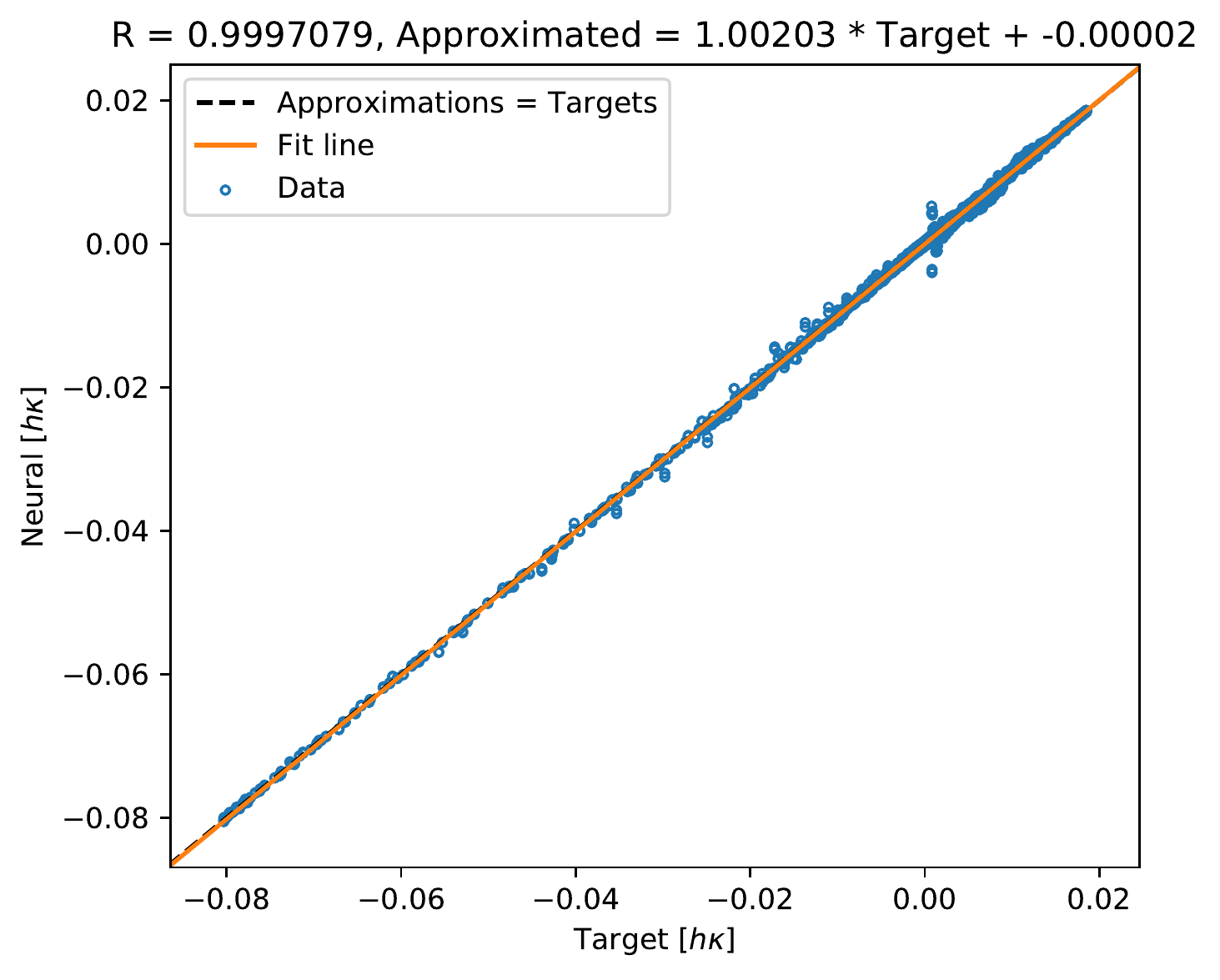}
		\caption{\footnotesize Using standardization}
		\label{fig:results.10.full.rose.a120.b305.nnet}
	\end{subfigure}
	
	\caption{Comparison of the generalization behavior of the network-only model for $\nu = 10$ when inferring $h\kappa$ along $\Gamma_2$.  (Color online.)}
	\label{fig:pathological.10.full.rose.a120.b305.comparison}
\end{figure}

As a prelude to this accuracy convergence study, we begin by analyzing the architectural complexity of $F_{net}^{(\nu)}$ and $F_{hyb}^{(\nu)}$.  Table \ref{tbl:results.convergence.topology} shows the relation among mesh size, number of trainable parameters, and number of computing units in each of the models' four hidden layers.  It also includes the cardinality of $\mathcal{D}^{(\nu)}$ and a few curvature error statistics.

There are a few points in Table \ref{tbl:results.convergence.topology} worth discussing.  As noted above, network-only models must account for the full curvature spectrum; thus, $F_{net}^{(\nu)}$ requires exposure to both positive and negative curvature samples.  Some direct consequences of this are longer training times and more complex architectures than $F_{hyb}^{(\nu)}$.  

A second observation drawn from Table \ref{tbl:results.convergence.topology} concerns the lack of correlation between the minimum cell width and the size of the hidden layers in $F_{net}^{(\nu)}$.  We attribute such a disassociation to the need for sufficient neural capacity to keep up with the increasing variety of sample patterns as $h \rightarrow 0$.  Also, deciding on the hidden layer size is mostly arbitrary; frequently, we settled on the shown architectures after several trials.  Even though we seek to simplify all neural topologies as much as possible, it turned out to be easier to design $F_{hyb}^{(\nu)}$ than tweaking the hyperparameters for $F_{net}^{(\nu)}$.

Lastly, we reflect on the seemingly amiss statistics for $\nu = 10$ in Table \ref{tbl:results.convergence.topology}.  All neural networks up to $\nu = 9$ used PCA and whitening to transform the input vector $\vv{\phi}$, as described in Section \ref{subsec:TechnicalAspects}.  However, when we optimized $F_{net}^{(10)}$ with the same input transformation, the generalization performance on $\Gamma_2$ degraded significantly along regions where $h\kappa \sim 0$ (see Figure \ref{fig:pathological.10.full.rose.a120.b305.nnet}).  After further investigation, we found that typical standardization\footnote{Preprocessing inputs with SciKit-Learn's {\tt StandardScaler} object.} helped produce more accurate curvature estimations for stencils not included in $\mathcal{D}^{(10)}$ (see Figure \ref{fig:results.10.full.rose.a120.b305.nnet}).  There is still no sound explanation for this phenomenon, but overfitting probably caused such pathological behavior.  Fast training convergence due to whitening might be exacerbating this issue too.  For these reasons, we have incorporated $F_{net}^{(10)}$ optimized for normalized level-set values, similar to the usage reported in \cite{LALariosFGibou;LSCurvatureML;2021}.

\begin{table}[!b]
	\centering
	\small
	\bgroup
	\def\arraystretch{1.1}%
	\begin{tabular}{|l|l|c|r|c|c|c|c|c|}
		\hline
		$\nu$ & Approach & \makecell{Data set\\size $|\mathcal{D}^{(\nu)}|$} & Params. & \makecell{Units per\\hidden layer} & \makecell{Training mean\\squared error} & \makecell{Testing mean\\squared error} & \makecell{Testing mean\\absolute error} & \makecell{Training\\epochs} \\
		\hline \hline
		\multirow{2}{*}{7} & Network-only &  4'442,728 & 200,193 & 256 & $\eten{6.72215}{-2}$ & $\eten{7.86289}{-2}$ & $\eten{1.68835}{-1}$ & 446 \\
 		                   & Hybrid       &  2'330,785 &  99,721 & 180 & $\eten{5.39347}{-2}$ & $\eten{6.64149}{-2}$ & $\eten{1.57649}{-1}$ & 434 \\
		\hline \hline
		\multirow{2}{*}{8} & Network-only &  7'258,628 & 159,145 & 228 & $\eten{1.40125}{-2}$ & $\eten{1.37766}{-2}$ & $\eten{7.65332}{-2}$ & 609 \\
 		                   & Hybrid       &  3'629,297 &  89,081 & 170 & $\eten{1.17456}{-2}$ & $\eten{1.18338}{-2}$ & $\eten{7.07474}{-2}$ & 432 \\
		\hline \hline
		\multirow{2}{*}{9} & Network-only &  9'315,620 & 148,281 & 220 & $\eten{5.06425}{-3}$ & $\eten{4.52365}{-3}$ & $\eten{4.53981}{-2}$ & 602 \\
 		                   & Hybrid       &  5,205,788 &  52,521 & 130 & $\eten{6.00609}{-3}$ & $\eten{5.54471}{-3}$ & $\eten{5.11820}{-2}$ & 400 \\
		\hline \hline
		\multirow{2}{*}{10}& Network-only & 12'592,164 & 159,145 & 228 & $\eten{1.78543}{-2}$ & $\eten{1.09794}{-2}$ & $\eten{7.30014}{-2}$ & 261 \\
 		                   & Hybrid       &  6,273,339 &  31,401 & 100 & $\eten{3.36056}{-3}$ & $\eten{3.04211}{-3}$ & $\eten{4.05121}{-2}$ & 342 \\
		\hline
	\end{tabular}
	\egroup
	\caption{Topological analysis and learning and error statistics for network-only and hybrid inference systems.}
	\label{tbl:results.convergence.topology}
\end{table}

Next, we assess the quality of the mean curvature approximations along $\Gamma_2$ at various resolutions.  As in Section \ref{subsec:EvaluationAtLowResolution}, we start by discretizing the computational domain with an adaptive grid.  The $1\times 1$ domain macromesh comprises a single quadtree with a maximum level of refinement of $\ell_{max}^{\nu} = \nu$.  Also, we note that both $F_{hyb}^{(\nu)}$ and $F_{net}^{(\nu)}$ operate only on level-set functions reinitialized with ten iterations.  For reference, we include results from using 20 iterations, but just with the compound numerical approach.

\begin{figure}[t]
	\centering
	\begin{subfigure}[b]{0.3\textwidth}
		\includegraphics[width=\textwidth]{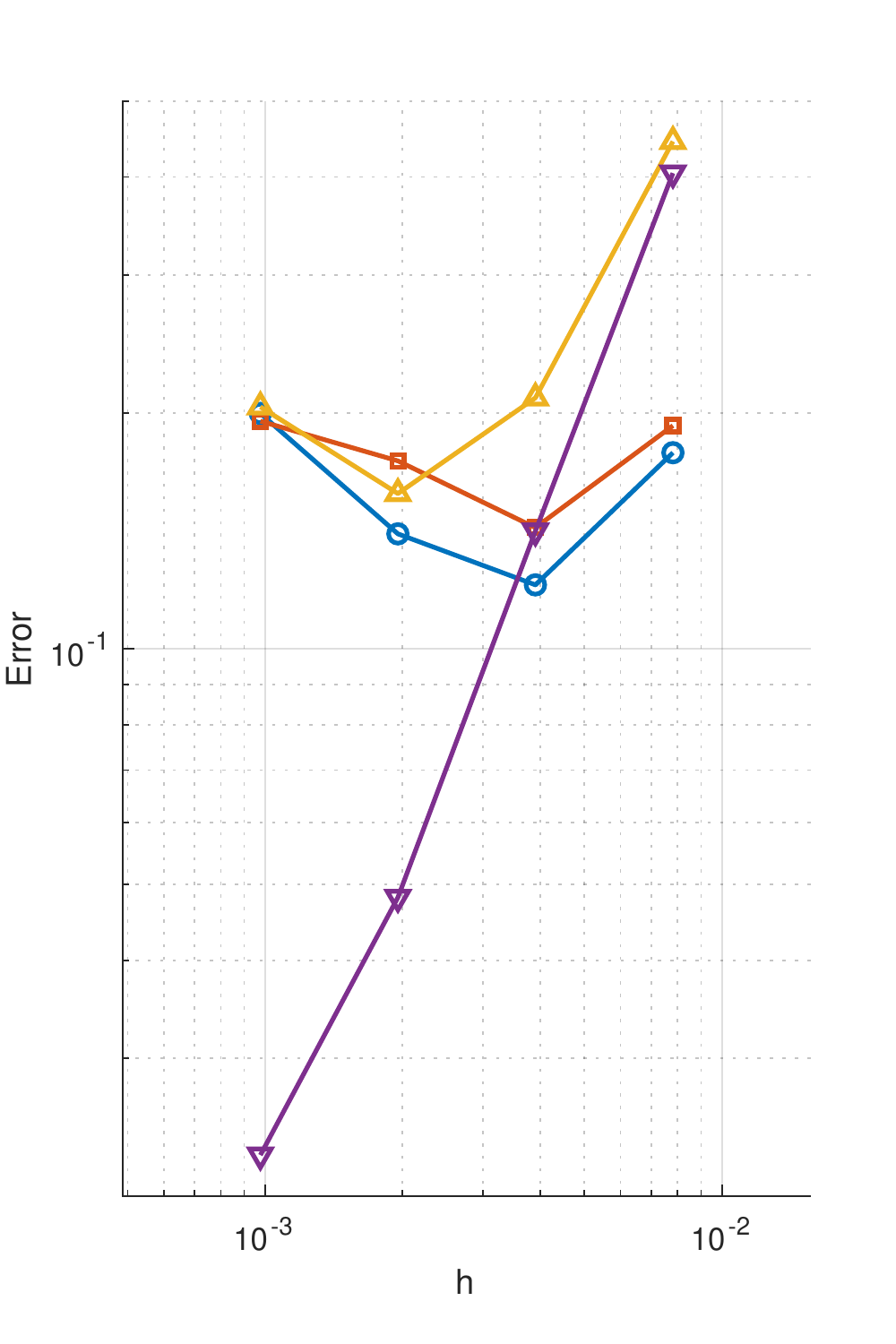}
        \caption{\footnotesize Mean absolute error}
        \label{fig:results.convergence.accuracy.mae}
    \end{subfigure}
    ~
	\begin{subfigure}[b]{0.3\textwidth}
		\includegraphics[width=\textwidth]{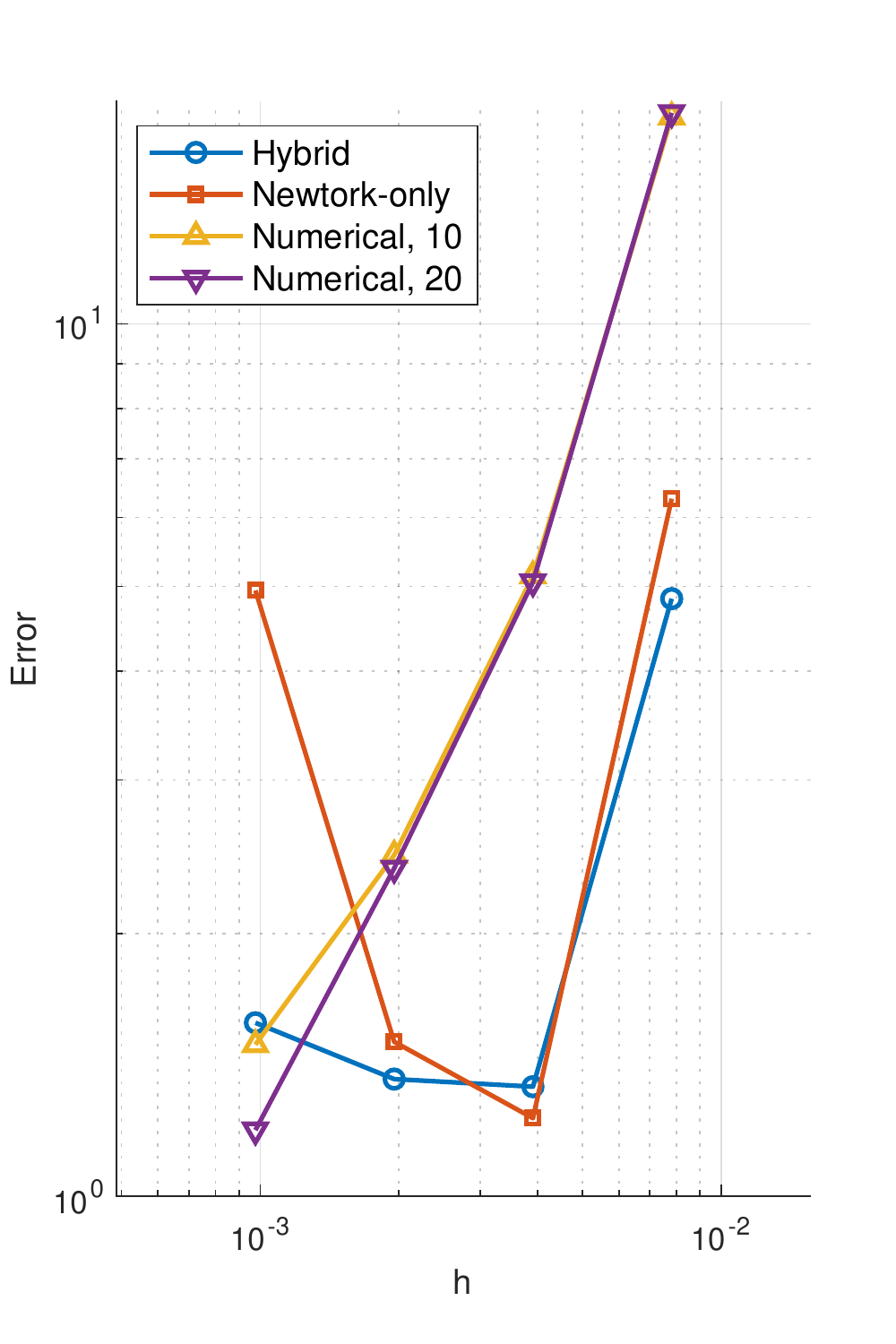}
		\caption{\footnotesize Maximum absolute error}
		\label{fig:results.convergence.accuracy.maxae}
	\end{subfigure}
	   
	\caption{Curvature estimation accuracy in the (a) $L^1$ and (b) $L^\infty$ norms using the proposed hybrid inference system, a network-only framework as in \cite{CurvatureML19, LALariosFGibou;LSCurvatureML;2021, VOFCurvature3DML19}, and the compound numerical method with 10 and 20 iterations to reinitialize $\phi_{rose}$.  (Color online.)}
	\label{fig:results.convergence.accuracy}
\end{figure}

Figure \ref{fig:results.convergence.accuracy} summarizes the curvature estimation convergence of the proposed hybrid inference system.  Tables \ref{tbl:results.convergence.mae} and \ref{tbl:results.convergence.maxae}, along with Table \ref{tbl:results.7.hybrid.rose.a120.b305.summary}, provide a more detailed $\kappa$ error analysis.  When contrasted with the compound numerical method, our approach is superior in the error $L^\infty$ norm, particularly in low resolutions, i.e., $\nu = 7, 8, 9$.  The hybrid inference system even outperforms $G_h(\cdot)$ when we employ twice the number of operations to reinitialize the level-set function.  Regarding average correctness, the hybrid algorithm is also more accurate than, if not comparable to, the numerical method whenever $\phi_{rose}$ is reinitialized with ten iterations.  In like manner, Figure \ref{fig:results.convergence.accuracy.mae} demonstrates that the 10-iteration-based machine learning and numerical approaches converge to the same accuracy as the mesh size decreases.  But, when one uses 20 operations for redistancing and $h \leqslant 2^{-9}$, $G_h(\cdot)$'s precision is higher in the $L^1$ norm.  This expected behavior is a consequence of more regions becoming well-resolved, which are indubitably the patterns where the numerical method excels.  In this regard, the end user must plan ahead and make an informed decision on how many reinitialization steps to use for training.  Due to the lack of physical constraints, our curvature networks are not guaranteed to extrapolate accurately beyond the quality of the sampled level-set field observed during optimization.

\begin{table}[!b]
	\centering
	\small
	\bgroup
	\def\arraystretch{1.1}%
	\begin{tabular}{|r|c|r|c|r|c|r|c|r|}
		\hline
		$\nu$ & Hybrid & Order & Network-only & Order & \makecell{Numerical\\10 iters.} & Order & \makecell{Numerical\\20 iters.} & Order \\
		\hline \hline
		8     & 0.120624 &  0.56 & 0.142708 &  0.43 & 0.209020 &  1.09 & 0.141279 & 1.52 \\
		\hline
		9     & 0.140138 & -0.22 & 0.173657 & -0.28 & 0.157754 &  0.41 & 0.048133 & 1.55 \\
		\hline
		10    & 0.199658 & -0.51 & 0.195176 & -0.17 & 0.203543 & -0.37 & 0.022570 & 1.09 \\
		\hline
	\end{tabular}
	\egroup
	\caption{Convergence analysis with respect to $\kappa$'s mean absolute error ($L^1$ norm) in $\Gamma_2$ at different resolutions.}
	\label{tbl:results.convergence.mae}
\end{table}

\begin{table}[!b]
	\centering
	\small
	\bgroup
	\def\arraystretch{1.1}%
	\begin{tabular}{|r|c|r|c|r|c|r|c|r|}
		\hline
		$\nu$ & Hybrid & Order & Network-only & Order & \makecell{Numerical\\10 iters.} & Order & \makecell{Numerical\\20 iters.} & Order \\
		\hline \hline
		8     & 1.335203 &  1.86 & 1.228807 &  2.36 & 5.141835 & 1.75 & 5.059004 & 1.79 \\
		\hline
		9     & 1.362297 & -0.03 & 1.503125 & -0.29 & 2.460210 & 1.06 & 2.375857 & 1.09 \\
		\hline
		10    & 1.580569 & -0.21 & 4.947860 & -1.72 & 1.492015 & 0.72 & 1.190646 & 1.00 \\
		\hline
	\end{tabular}
	\egroup
	\caption{Convergence analysis with respect to $\kappa$'s maximum absolute error ($L^\infty$ norm) in $\Gamma_2$ at different resolutions.}
	\label{tbl:results.convergence.maxae}
\end{table}

Figure \ref{fig:results.convergence.accuracy} further shows that our proposed strategy is more accurate than $F_{net}^{(\nu)}$ in nearly all tested resolutions in both error metrics.  By combining these results with the information in Table \ref{tbl:results.convergence.topology}, it is evident that the hybrid solver is more efficient than the network-only inference system.  Indeed, the main advantage of the hybrid mechanism is that it needs much fewer computing units to attain the same or better accuracy than the full curvature model.

Unlike the compound numerical method, none of the neural strategies tested here exhibits a definite convergence order.  Qi \etal have corroborated this datum in their study of machine learning and VOF \cite{CurvatureML19}.  In particular, our experiments show that it gets increasingly complicated and quite resource-intensive to train either $F_{net}^{(\nu)}$ or $F_{hyb}^{(\nu)}$ as $h_{\nu} \rightarrow 0$.  Even with enormous data sets $\mathcal{D}^{(\nu)}$, there is a latent risk of overfitting, as seen in Figure \ref{fig:pathological.10.full.rose.a120.b305.nnet}.  Overall, optimizing neural models to estimate curvature in the level-set method is not a straightforward process for extremely small grid spacings.  In that case, it is rather convenient to rely on traditional numerical schemes; they are less expensive and deliver more accurate and predictable approximations than curvature solvers based on neural models.


\colorsubsection{Circular-interface-based convergence study}
\label{subsec:CircularConvergenceStudy}

We close this section with a second convergence analysis based on circular interfaces.  The goal is to demonstrate our neural models' ability to estimate curvature at a comparable level of accuracy to state-of-the-art technologies in other implicit representations.  In this case study, we contrast the response of our $F_{hyb}^{(\nu)}$ functions introduced in Section \ref{subsec:FlowerShapeConvergenceStudy} with the hybrid particle-VOF method of Karnakov \etal \cite{Karnakov;etal;HybridParticleMthdVOFCuravture;2020}.  Their novel algorithm relies on fitting circular arcs to a piecewise linear interface reconstruction and computes curvature with the help of strings of particles in equilibrium.  The authors have shown in several benchmark problems that such an algorithm can significantly improve accuracy at low and sub-scale resolutions.  Moreover, combining this particle method with generalized height functions \cite{Popinet:09:An-accurate-adaptive} has proven more effective than blending standard height functions \cite{Cummins;etal;HeightFunctions;2005} and parabolic fitting \cite{Renardy;Renardy:02:PROST:-A-Parabolic-R}.

We replicate in this experiment the curvature problem described in \cite{Karnakov;etal;HybridParticleMthdVOFCuravture;2020} (refer to Section \href{https://www.sciencedirect.com/science/article/pii/S0301932219306615\#sec0017}{3.1}).  The test entails measuring the error $L^2$ and $L^\infty$ norms as the number of cells to discretize a circle of radius $R$ increases.  To this end, we vary the ratio $R/h_\nu$ for a fixed $R = 2/128$ and $h_\nu = 2^{-\nu}$, where $\nu = 7,8,9,10$.  Here, we have chosen $2/128$ because \textit{none} of our models has been specifically trained with circles having such an $R$.  Also, this radius defines almost the smallest interface we can resolve with the coarsest-grid model trained for $h = 2^{-7}$.

The experiment involves collecting samples separately for each of the four $\nu$ values by drawing $100$ centers $\vv{x}_0 \in \left[-h_\nu/2, +h_\nu/2\right]^2$ from a random uniform distribution.  In particular, we are only interested in reinitialized level-set fields defined by $\phi_{rls}^c(x,y)$ in \eqref{eq:circularInterfaces}.  Then, we quantify the curvature error over the sample set $\mathcal{S}_\nu$ by computing the relative norms

\begin{equation}
\tilde{L}_\nu^2 = \left(\frac{1}{\left|\mathcal{S}_\nu\right|}\sum_{s\in \mathcal{S_\nu}} \left(\frac{\kappa_s - \kappa^*}{\kappa^*}\right)^2 \right)^{1/2} \quad \textrm{and} \quad
\tilde{L}_\nu^\infty = \max_{s\in \mathcal{S_\nu}} \left|\frac{\kappa_s - \kappa^*}{\kappa^*}\right|,
\label{eq:RelativeNorms}
\end{equation}
where $\kappa^*$ is the exact value and $s$ is a sampled grid point next to $\Gamma$.  These equations are adaptations to expressions (14) and (15) in \cite{Karnakov;etal;HybridParticleMthdVOFCuravture;2020}.  In our case, the $\tilde{L}_\nu^2$ and $\tilde{L}_\nu^\infty$ metrics above have resulted solely from evaluating $F_{hyb}^{(\nu)}$ since $1/R = \kappa = 64 > \kappa_{flat}^{(\nu)}$ for all $\nu$ (see Section \ref{subsec:FlowerShapeConvergenceStudy} and line \ref{alg:hybridApproach:condition} in Algorithm \ref{alg:hybridApproach}).

\begin{figure}[t]
	\centering
	\begin{subfigure}[b]{0.3\textwidth}
		\includegraphics[width=\textwidth]{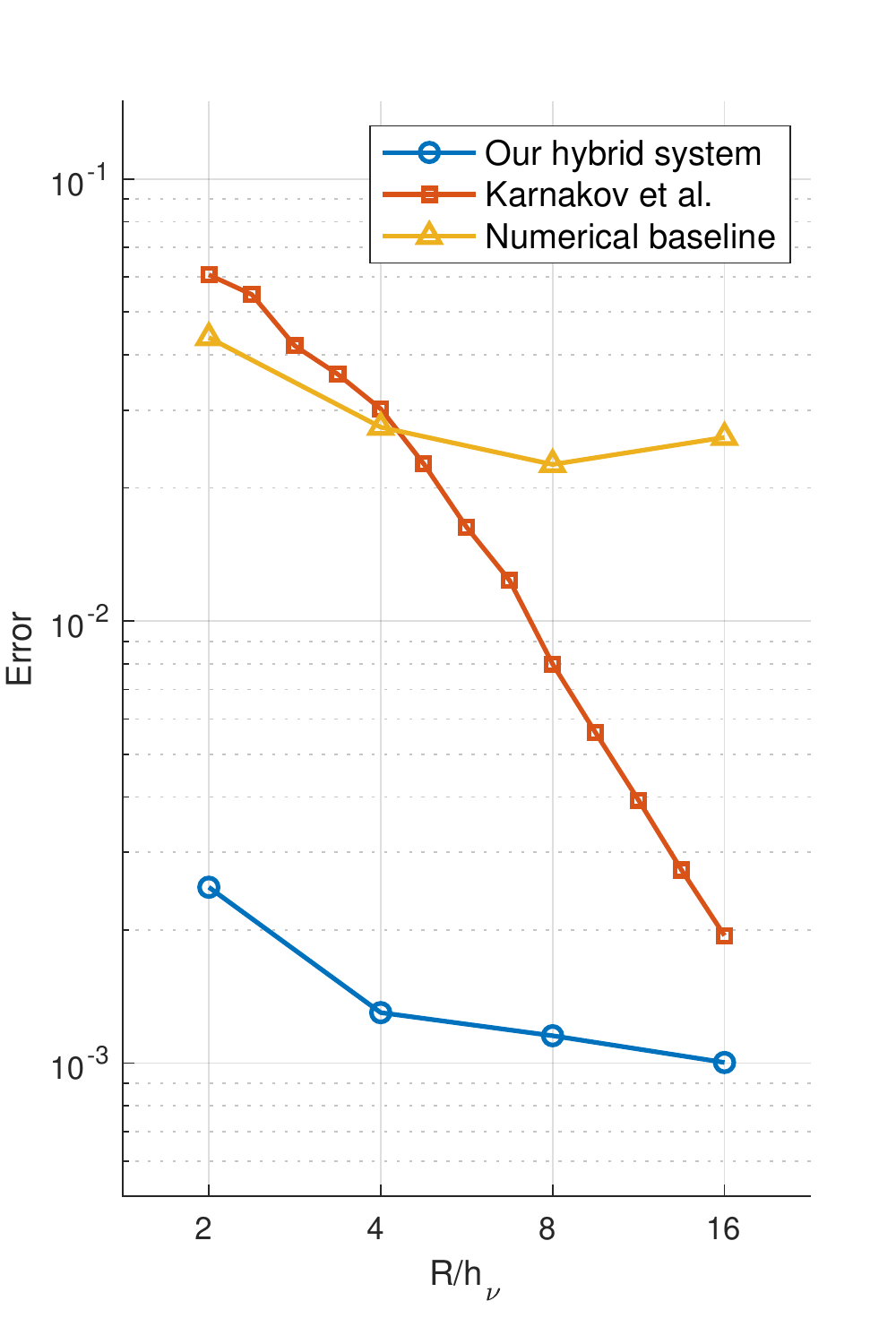}
        \caption{\footnotesize $\tilde{L}_\nu^2$}
        \label{fig:results.convergence.circle.l2}
    \end{subfigure}
    ~
	\begin{subfigure}[b]{0.3\textwidth}
		\includegraphics[width=\textwidth]{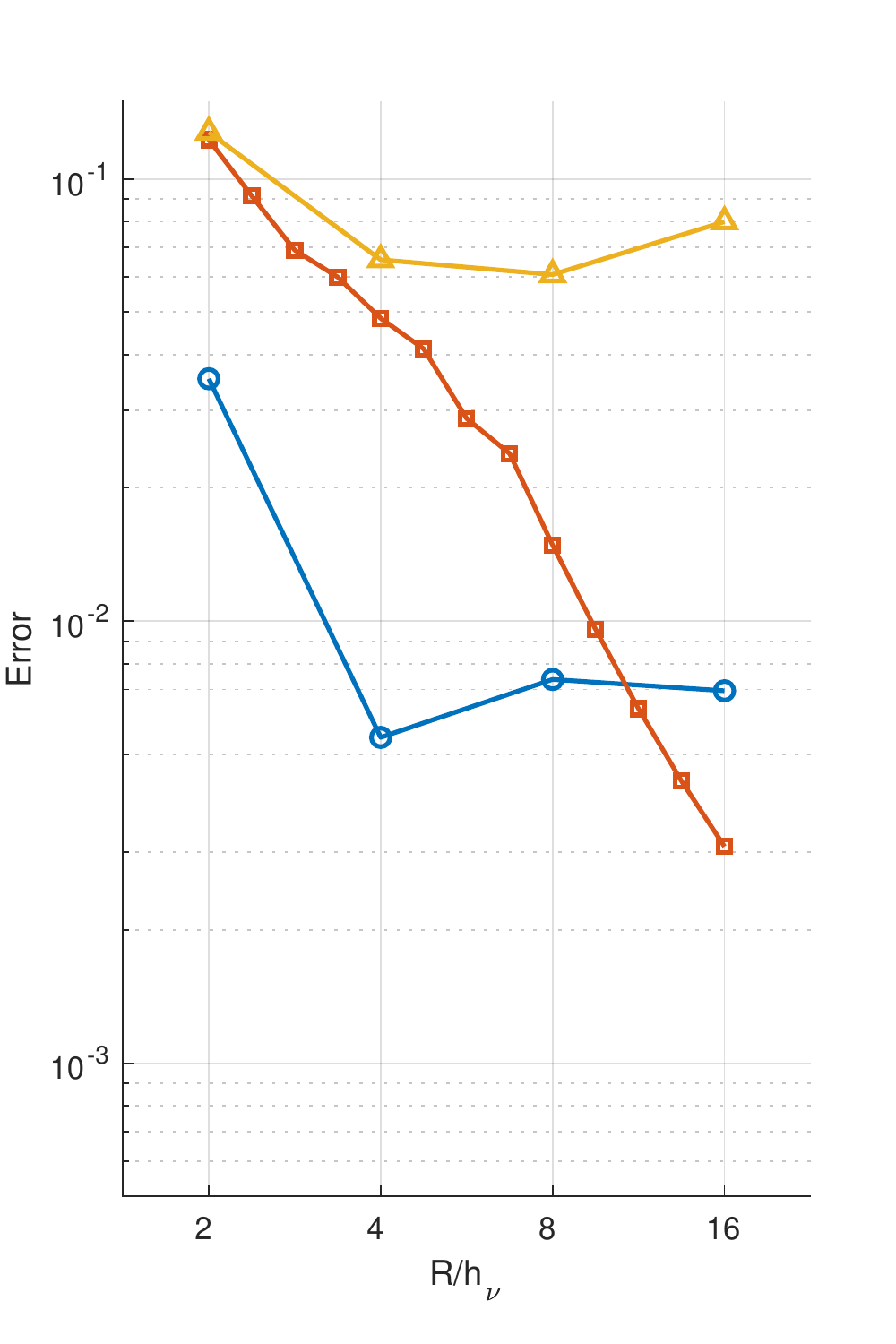}
		\caption{\footnotesize $\tilde{L}_\nu^\infty$}
		\label{fig:results.convergence.circle.linfty}
	\end{subfigure}
	   
	\caption{Circular-interface relative curvature error for $\nu = 7,8,9,10$ using the proposed $F_{hyb}^{(\nu)}$ neural networks and the numerical baseline with 10 steps for redistancing $\phi_{rls}^c(x,y)$.  Also, we show the results for the combined particle and generalized height function method of \cite{Karnakov;etal;HybridParticleMthdVOFCuravture;2020}.  (Color online.)}
	\label{fig:results.convergence.circle}
\end{figure}

Figure \ref{fig:results.convergence.circle} contrasts our system's relative curvature error with the hybrid method in \cite{Karnakov;etal;HybridParticleMthdVOFCuravture;2020} (see their Figure \href{https://www.sciencedirect.com/science/article/pii/S0301932219306615\#fig0010}{10}).  To ease the comparison, we have reproduced Karnakov and coauthors' results for the mesh-size range evaluated in this test.  A swift analysis of Figure \ref{fig:results.convergence.circle} reveals that our models are superior to the hybrid particle-VOF approach around $R/h_\nu = 2, 4, 8$ (i.e., for $\nu=7,8,9$).  However, unlike their framework, there is no neural network inference convergence, which corroborates our findings in Section \ref{subsec:FlowerShapeConvergenceStudy}.  Clearly, higher-order schemes are more precise and appropriate than neural models in fine grids; nevertheless, we have shown that data-driven solvers bear an unparalleled potential to address under-resolution numerical problems and their undermining effects.


\colorsection{Conclusions}
\label{sec:Conclusions}

We have presented a hybrid strategy for estimating curvature in the level-set method using machine learning\footnote{Our neural networks and preprocessing modules are available at \url{https://github.com/UCSB-CASL/HybridDLCurvature}.}.  Our hybrid inference system substantially enhances the concept of the curvature regression models introduced in \cite{LALariosFGibou;LSCurvatureML;2021}.  To increase accuracy, we fit dedicated neural networks to data sets constructed from sinusoidal- and circular-interface samples.  The core of our framework is a switching mechanism that relies on conventional numerical schemes to gauge curvature.  If the curvature magnitude is larger than a resolution-dependent threshold, it uses a preprocessing module and a neural component to deliver a better $h\kappa$ approximation.  Our networks do not need to be trained on the whole curvature spectrum, leading to compact architectures, improved data exploitation, and faster parameter adaptation.

Our experiments have shown that the hybrid strategy is more accurate than the compound numerical method and more efficient than full-curvature-spectrum neural networks.  In particular, the hybrid inference system yields remarkable gains in precision when dealing with coarse grids and when the interfaces exhibit pronounced concave or convex regions. Similarly, it can be applied to uniform grids and quad-/oc-trees \cite{Min;Gibou:07:A-second-order-accur} and within the local level-set framework \cite{Peng;Merriman;Osher;etal:99:A-PDE-based-fast-loc, Nielsen;Museth:06:Dynamic-Tubular-Grid, Brun;Guittet;Gibou:12:A-local-level-set-me}.  However, we have observed some deterioration in neural correctness as resolution increases.  This unexpected result has been consistent in both the hybrid and the compatible network-only model, as reported in Section \ref{subsec:FlowerShapeConvergenceStudy}.  We think the issue is probably a consequence of overfitting, which gets exacerbated by using PCA and whitening to treat data inputs.  Thus, further research on this pathologic behavior is required to understand how to design adequate data sets for training in high resolutions.  Likewise, we should note our strategy's limitation to handle only smooth interfaces.  Compared to the local level-set extraction method \cite{Ervik;Lervag;Munkejord;LOLEX;2014} and Macklin and Lowengrub's curve-fitting approach \cite{Macklin;Lowengrub:05:Evolving-interfaces-, Macklin;Lowengrub;ImprovedCurvatureAppTumorGrowth;2006}, for example, our neural networks cannot tackle coalescence or bodies folding back onto themselves.  Also, after verifying the results of the recent hybrid particle method of Karnakov \etal \cite{Karnakov;etal;HybridParticleMthdVOFCuravture;2020}, we have identified a potential niche for neural curvature estimation at sub-scale (i.e., for radii below $1.5h$).  In this regard, we leave to future endeavors the study of these circumstances to increase the appeal of our data-driven curvature solver.

The hybrid inference system and its corresponding neural network are part of our long-term plan to develop a family of level-set machine learning methods.  With this goal in mind, we have reserved the coupling of the hybrid strategy with a full-fledged simulation for upcoming work to assess the response of our regression models to advection processes.  Also, a subordinate task to such a simulation will require the migration of the inference system from TensorFlow/Keras to C++.  Despr\'{e}s and Jourdren have already done so with their VOF machine learning schemes in \cite{DespresJourdren;MLDesignOfVOF;20} with the aid of \verb|keras2cpp| \cite{keras2cpp;20}.  We believe the same approach is promising for future developments, alongside network pruning to reduce evaluation times \cite{ZG18, frankle2019lottery}.


\color{aqua}\sffamily\section*{Acknowledgements}\rmfamily\color{black}

Use was made of computational facilities purchased with funds from the National Science Foundation (CNS-1725797) and administered by the Center for Scientific Computing (CSC).  The CSC is supported by the California NanoSystems Institute and the Materials Research Science and Engineering Center (MRSEC; NSF DMR 1720256) at UC Santa Barbara.



{\footnotesize
\biboptions{sort&compress}
\bibliographystyle{unsrt}
\bibliography{references}}

\end{document}